\documentclass[lettersize,journal]{IEEEtran}
\usepackage{amsmath,amsfonts}
\usepackage{algorithmic}
\usepackage{algorithm}
\usepackage{array}
\usepackage{booktabs}
\usepackage[caption=false,font=normalsize,labelfont=sf,textfont=sf]{subfig}
\usepackage{textcomp}
\usepackage{stfloats}
\usepackage{url}
\usepackage{verbatim}
\usepackage{graphicx}
\usepackage{color}
\usepackage{multirow}

\def\BibTeX{{\rm B\kern-.05em{\sc i\kern-.025em b}\kern-.08em
    T\kern-.1667em\lower.7ex\hbox{E}\kern-.125emX}}
\usepackage{balance}
\begin{document}
\title{GLiRA: Black-Box Membership Inference Attack via Knowledge Distillation}

\author{Andrey V. Galichin, Mikhail Pautov, Alexey Zhavoronkin, Oleg Y. Rogov and Ivan Oseledets
\thanks{M. Pautov, O. Y. Rogov and I. Oseledets are with the Artificial Intelligence Research Institute, Moscow, Russia.}
\thanks{A. V. Galichin, M. Pautov, O. Y. Rogov and I. Oseledets are with the Skolkovo Institute of Science and Technology, Moscow, Russia.}
\thanks{A. Zhavoronkin is with the Moscow Institute of Physics and Technology, Dolgoprudny, Russia.}
\thanks{M. Pautov is with the ISP RAS Research Center for Trusted Artificial Intelligence, Moscow, Russia.}
}


\markboth{
}%
{GLiRA: Black-Box Membership Inference Attack via Knowledge Distillation}

\maketitle

\begin{abstract}
While Deep Neural Networks (DNNs) have demonstrated remarkable performance in tasks related to perception and control, there are still several unresolved concerns regarding the privacy of their training data, particularly in the context of vulnerability to Membership Inference Attacks (MIAs). In this paper, we explore a connection between the susceptibility to membership inference attacks and the vulnerability to distillation-based functionality stealing attacks. In particular, we propose {GLiRA}, a distillation-guided approach to membership inference attack on the black-box neural network. We observe that the knowledge distillation significantly improves the efficiency of likelihood ratio of membership inference attack, especially in the black-box setting, i.e., when the architecture of the target model is unknown to the attacker. 
We evaluate the proposed method across multiple image classification datasets and models and demonstrate that likelihood ratio attacks when guided by the knowledge distillation, outperform the current state-of-the-art membership inference attacks in the black-box setting.  
\end{abstract}


\section{Introduction}

Recent studies \cite{carlini2019secret,zhang2021understanding,wu2023membership} have shown that machine learning models are used to memorize the training data and, hence, have a vast spectrum of data-driven vulnerabilities, such as membership inference attacks \cite{shokri2017membership,sablayrolles2019white,choquette2021label,hu2022membership,rahman2018membership,carlini2022membership,DBLP:conf/iclr/WenBKBGGG23,LZBZ22,TEAR2023}.  
Informally, a membership inference attack is a procedure to determine whether a particular data sample was a part of the training dataset of the given target neural network. Given the data sample $(x,y)$, 
the majority of membership inference attacks (\cite{carlini2022membership,watson2021importance,dubinski2024towards,ye2022enhanced,bertran2024scalable}) are based on designing a specific statistic $s((x,y))$ to distinguish between the models trained on a sample $(x,y)$ and those which were not. In other words, given $\mathcal{H}_1$ as the class of models which have $(x,y)$ in their training set, and $\mathcal{H}_2$ as the class of models which did not see $(x,y)$ during training, the statistics $s\left((x,y) \vert \mathcal{H}_1\right)$ and $s\left((x,y) \vert \mathcal{H}_2\right)$ are assumed to have substantially different distributions. If the difference is significant, these statistics can then be used to determine to which class the target model belongs, $\mathcal{H}_1$ or $\mathcal{H}_2$.


The efficiency of such membership inference attack methods depends on the design of classes $\mathcal{H}_1$ and $\mathcal{H}_2$. The most widespread approach to design $\mathcal{H}_1$ and $\mathcal{H}_2$ is to train \emph{shadow models} \cite{shokri2017membership,sablayrolles2019white,carlini2022membership,DBLP:conf/iclr/WenBKBGGG23}.  Usually, shadow models are of the same architecture as the target model $f$, and they 
are trained on the random data sampled from the same distribution $\mathcal{D}$, which was used to train the target model $f$. These shadow models are then used to compute the values of the test statistics $s((x,y) \vert \mathcal{H}_1)$ and $s((x,y) \vert \mathcal{H}_2)$, so they have to be sufficiently different from each other. To estimate the densities of the test statistics, an attacker has to train a lot of shadow models (in the work of \cite{carlini2022membership}, between $64$ and $256$ were used). More than that, it is known (\cite{carlini2022membership,bertran2024scalable}) that the shadow models should be of the same or more complicated architecture than the target model to ensure high efficiency of the membership inference attack. Since the architecture of the target model is often unknown, it makes membership inference attacks barely feasible in practice. 


In this work, we focus on membership inference attacks in the setting of a classification problem within the image domain. We study the effect of the knowledge distillation \cite{hinton2015distilling} on the success of the membership inference attack on a black-box neural network. Namely, we hypothesize that training the shadow models via knowledge distillation of the target model significantly improves the precision of the membership inference attack, especially in the black-box settings, i.e., when the architecture of the target model is unknown. We propose \emph{GLiRA}, or \textbf{G}uided \textbf{Li}kelihood \textbf{R}atio \textbf{A}ttack, a novel approach to conduct a membership inference attack without knowledge about the target model's architecture and training dataset. We train shadow models on different subsets of the hold-out dataset, resembling the offline membership inference attack \cite{carlini2022membership}. To ensure a higher degree of similarity with the target model, each shadow model is trained via knowledge distillation.  

The contribution of this paper is the following:
\begin{itemize}
    \item We propose \emph{GLiRA}, a novel membership inference attack method based on knowledge distillation of the target model. The proposed approach does not require excessive information about the target model, such as its architecture or training dataset. 
    \item We adapt the knowledge distillation to ensure that the shadow models learn the underlying distribution of the logits of the target model. 
    \item We evaluate our approach on several image classification datasets in different experimental setups and show that \emph{GLiRA} outperforms the state-of-the-art membership inference attacks in the majority of the settings considered, especially when no excessive information about the target model is known. 
\end{itemize}

\section{Background}

\subsection{Machine Learning and Classification Problem}

In Machine Learning, we conceptualize a classification neural network as a function mapping inputs to a set of probabilities across different (usually predefined) classes. Here we denote it as $f_{\theta}: \mathcal{X} \rightarrow [0, 1]^K$, with $f_{\theta}(x)_y$ representing the probability to assign object $x \in \mathcal{X}$ to class $y \in [1,\dots,K]$. Suppose we have a dataset $D$, which is a sample from a broader distribution $\mathbb{D}$. We use $f_{\theta} \leftarrow \mathcal{T}(D)$ to signify the process of training a neural network $f$ parameterized by weights $\theta$ with a training algorithm $\mathcal{T}$ on $D$. In our setting, the training of the network is represented by a sequence of stochastic gradient descent steps to optimize a predefined loss function $l$:
\begin{equation}
    \theta_{i+1} \leftarrow \theta_i - \eta\sum\limits_{(x,y) \in B}\nabla_{\theta}l(f_{\theta_i}(x), y),
\end{equation}
where $B \subset D$ refers to a subset of the training data (mini-batch) and $\eta > 0$ is the learning rate. In this work, we use the cross-entropy loss function as one of the most suitable for the classification problem:
\begin{equation}
    l(f_{\theta}(x), y) = -\log(f_{\theta}(x)_y).
\end{equation}
The prediction $f(x)$ of the neural network $f$ can be represented as $f(x) = \sigma(z(x))$ (we will omit $\theta$ when referring to the neural network function is implicit from the context), where $z: \mathcal{X} \to \mathbb{R}^K$ is a mapping to feature outputs (i.e. logits), and $\sigma(z) = (\sigma(z)_1, \dots, \sigma(z)_K)$ is the softmax function that converts these outputs into probabilities: 
\begin{equation}
    \sigma(z)_i = \frac{\exp(z_i)}{\sum\limits_{j=1}^{K}\exp(z_j)}.
\end{equation}
\subsection{Membership Inference Attacks}


The objective of the membership inference attack (MIA) \cite{shokri2017membership} is to determine whether a specific data sample was presented in the training data of the target model or no. MIAs demonstrate that, under mild assumptions about the target model, it is possible to identify a part of its training dataset, leading to possible leakage of private data. To broaden the scope of practical applications of neural networks, it is important to have a reliable tool to assess their vulnerability to the leakage of private training data. 



Formally, given a data sample $(x, y)$, the target model $f_{\theta}$ trained on (possibly fully unknown) dataset $D$ and additional information about $f_{\theta}$  denoted by $I$, membership inference attack $\mathcal{A}$ is  defined as the  function
\begin{equation}
    \mathcal{A}(x,f_{\theta},I) =
    \begin{cases}
        0, \ \text{if $x \notin D$}, \\
        1, \ \text{if $x \in D$}.
    \end{cases}
\end{equation}
A detailed explanation of the proposed membership inference attack will be given in the next sections.


\subsection{Knowledge Distillation}
\label{subsec:kd}


Knowledge distillation (KD) is a process of transferring knowledge from a large model to a smaller one without significant loss in the performance on the downstream task. Initially proposed in \cite{hinton2015distilling}, KD is represented as a framework where a smaller ``student'' network $f_s$ is trained to mimic the outputs of the larger ``teacher'' network $f_t$. The student does this by training not only on the ``hard'' labels of the training set but also on the ``soft`` probabilities produced by the teacher for each class. These soft probabilities carry additional information about the dataset, such as relationships between different classes.

Namely, given an input $x$ and the number of classes $K$, the teacher's output is the vector of $K$ class probabilities $$f_t(x) = p^t = (p_1^t, \dots p_K^t).$$ Similarly, the student's output is the vector $$f_s(x) = p^s = (p_1^s, \dots p_K^s).$$ To further guide the student model, a temperature parameter $\tau$ is introduced to control the ``softness'' of probabilities, making the output distribution smoother and revealing more information about the teacher's output structure. With this, the probability of each class is calculated by 
\begin{equation}
\label{eq:softmax_probs}
    p_k^t(\tau) = \frac{\exp(z_k^t / \tau)}{\sum_{j=1}^{K}\exp(z_j^t / \tau)}, \quad p_k^s(\tau) = \frac{\exp(z_k^s / \tau)}{\sum_{j=1}^{K}\exp(z_j^s / \tau)},
\end{equation}
where $z_j^t$ and $z_j^s$ are the logits produced by the teacher and student models, respectively.

During training, the student's loss function is a combination of the loss function for the downstream task (cross-entropy loss, in our case) and a distillation loss, which measures the difference between the outputs of the student and teacher. In the work of \cite{hinton2015distilling}, the Kullback-Leibler (KL) divergence was proposed as the distillation loss:

\begin{equation}
    \mathcal{L}_{\text{KL}}(p^s(\tau), p^t(\tau)) = \tau^2\sum_{j=1}^Kp_j^t(\tau)\log\frac{p_j^t(\tau)}{p_j^s(\tau)}.
\end{equation}
This function minimizes the difference between two probability distributions represented by $p^s$ and $p^t$. The student's loss function is then given by

\begin{equation}
\begin{split}
\label{eq:knowledge_distillation_kl}
    \mathcal{L}_{\text{KD}}(y, p^s(\tau), p^t(\tau)) = \ &\alpha\mathcal{L}_{\text{KL}}(p^s(\tau), p^t(\tau)) + \\ (1 - &\alpha)\mathcal{L}_{\text{CE}}(y, p^s(1)).    
\end{split}
\end{equation}
Here, $y$ represents the true label, $\mathcal{L}_{\text{CE}}$ is the cross-entropy loss, and $\alpha \in [0,1]$ is a hyperparameter that balances the two terms.

Lately, the authors of \cite{kim2021comparing} reported that, as $\tau$ increases, the KL divergence loss focuses on the \textit{logit matching} compared to \textit{label matching} when $\tau \rightarrow 0$. Also, they demonstrated that logit matching affects the performance of the student network positively. By analyzing the properties of KL loss, the work showed that there exists a relationship between KL and Mean-Squared-Error (MSE) computed between the logits of a student and teacher networks in the form 

\begin{equation}\label{eq:mse_loss}
\mathcal{L}_{\text{MSE}}(z^s, z^t) = ||z^s - z^t||^2_2,    
\end{equation}
when $\tau \rightarrow \infty$. Specifically, using MSE instead of KL as a distillation loss results in a more efficient \textit{logit matching}, leading to better distillation quality. In this scenario, the student's loss in the KD framework is calculated by

\begin{equation}
\begin{split}
\label{eq:knowledge_distillation_mse}
    \mathcal{L}_{\text{KD}}(y, z^s, z^t) = \ &\alpha\mathcal{L}_{\text{MSE}}(z^s, z^t) + \\ (1 - &\alpha)\mathcal{L}_{\text{CE}}(y, \sigma(z^s)).
\end{split}
\end{equation}

In this paper, we aim to explore the benefits of knowledge distillation approach for MIA, and show that 
\textit{logits matching} method using $\mathcal{L}_{\text{MSE}}$ instead of $\mathcal{L}_{\text{KL}}$ can help to further improve our MIA attack capabilities. 

\section{Related Work}

Membership inference attack (MIA) \cite{shokri2017membership} is the common name for the methods to determine whether or not a particular example was presented in the training dataset of the given target model. In the original work, the authors train multiple shadow models to mimic the behavior of the target model and use its outputs to train  an auxiliary classifier to predict the membership status of the data samples. 
The black-box methods for membership inference use the information about the loss of the target model \cite{yeom2018privacy,sablayrolles2019white,LZBZ22,ye2022enhanced}, labels \cite{choquette2021label,li2021membership}, or the functions of the loss value \cite{watson2021importance, carlini2022membership}. Among the other approaches, there are ones using quantile regression \cite{bertran2024scalable}, knowledge distillation \cite{ye2022enhanced,jagielski2024students}, neighbourhood comparison \cite{mattern2023membership} and parameter regularization \cite{tan2023blessing}.



In \cite{carlini2022membership}, the authors argue that a method to MIA as a classification problem should be evaluated by computing the values of true positive rate (TPR) at low values of false positive rate (FPR)  instead of classic average-case metrics (e.g., accuracy or the area under the ROC curve). They introduce Likelihood Ratio Attack (LiRA), an approach to membership inference attacks via statistical hypothesis testing. 
Namely, given the target model $f$ trained on an unknown dataset $D$ and a sample of interest $(x,y)$, they perform hypothesis testing, $H_0 \ \text{vs} \  H_1.$ Here $H_0: (x,y) \in D$ and $H_1: (x,y) \notin D$. Specifically, using the shadow models, they estimate the distribution of the models' confidence to assign $x$ to its ground truth class $y$. The authors improve their method by querying on multiple augmented versions of the data point $x$. 

After being shaped in \cite{hinton2015distilling}, knowledge distillation has been successfully integrated in different areas \cite{romero2015fitnets, zagoruyko2017paying, srinivas2018knowledge, park2019relational, heo2019comprehensive, pautov2024probabilistically}. However, in the field of membership inference attack, only the specific direction of knowledge distillation, called self-distillation \cite{furlanello2018born},  has been utilized so far. In this setting, the teacher model and student model(s) have identical architectures. In \cite{LZBZ22}, the authors note that most of the existing membership inference attack methods leverage only the information from the output of the given target model, relating these methods to the black-box ones. They propose to additionally exploit the information about the target model's training process during the membership inference attack. The authors do not deviate from the black-box setup of the attack: to integrate the information about the model' training and evaluate the data point's membership status based on the distilled models' behaviour at different distillation stages. The authors of \cite{ye2022enhanced} integrate the knowledge distillation into the process of the training of the shadow models. This work is the closest to ours, however it differs from ours in several important aspects. First of all, in our work, we additionally consider the true black-box setup, where an attacker is unaware of the architecture of the target model and, hence, can not adapt the knowledge distillation procedure accordingly. Secondly, we modify the loss function for the knowledge distillation term and experimentally show that it leads to the higher precision of the membership inference attack. Additionally, we evaluate the impact of the weight of the divergence term in the corresponding loss function on the success of the attack. 


\begin{figure}[ht!]
\centering
    \includegraphics[width=0.48\textwidth]{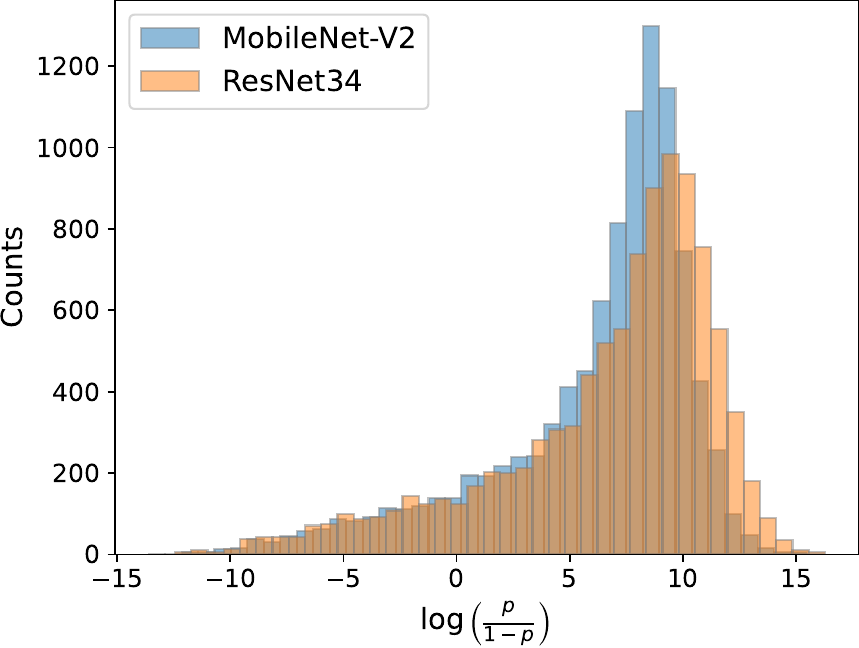}
    \caption{Histograms of the logits of the ground truth class for different architectures of the target model, CIFAR10 training dataset.  
    We observe a notable difference between the histograms, which can lead to a decreased alignment between target and shadow models if their architectures differ.}
    \label{fig:method_distribution}
\end{figure}

\section{Method}
\begin{figure*}
    \centering
    \includegraphics[width=1.0\textwidth]{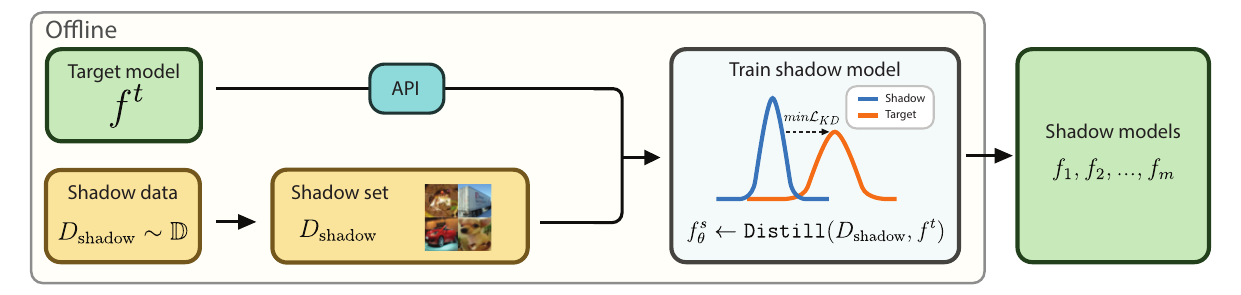}
    \caption{The illustration of the proposed pipeline for shadow models training. We are given the target model $f^t$, which can be queried via an API. We sample a training dataset $D_\text{shadow}$ from the underlying training data distribution $\mathbb{D}$, and train the shadow model using the knowledge distillation procedure $\texttt{Distill}(D_\text{shadow}, f^t)$. The process is repeated $m$ times to obtain the final set of shadow models $\{f_1, f_2, ..., f_m\}$. After that, an adversary can use the shadow models to determine the membership status of a given data point.}
    \label{fig:enter-label}
\end{figure*}
\subsection{Threat Model}

We follow a standard membership inference game as defined in \cite{carlini2022membership}. Namely, we introduce two parties participating in the game: a \textit{challenger} and an \textit{adversary}. The challenger samples a training dataset $D_t \leftarrow \mathbb{D}$, where $\mathbb{D}$ is the underlying training data distribution, and trains a model $f_t  \leftarrow \mathcal{T}(D_t)$. The adversary gets query access to the distribution $\mathbb{D}$ and to the model $f_t$. Given a point $(x^*, y^*) \leftarrow \mathbb{D}$, the attacker aims to determine whether $(x^*, y^*) \in \mathbb{D}$, or $(x^*, y^*) \notin \mathbb{D}$.

In this paper, we focus on a black-box scenario of MIAs, in which the adversary has access to the underlying training data distribution $\mathbb{D}$ (to train shadow models \cite{shokri2017membership,carlini2022membership}), and an output of the target model, which is a continuous vector of class probabilities. Our method does not require an adversary to know the architecture of the target model; still, our experiments show that the knowledge of the target model's architecture increases the efficiency of the membership inference attack, as expected.  

In our work, we apply Mean-Squared-Error Knowledge Distillation \cite{kim2021comparing}. To do so, we assume that the adversary is given either the output probabilities or the logits $z^t$ from the  Eq. \eqref{eq:softmax_probs}. 


\subsection{Knowledge Distillation for Shadow Model's Training}




Many Membership Inference Attacks require training shadow models to mimic the behaviour of the target model \cite{sablayrolles2019white, watson2021importance, carlini2022membership}. Usually, the shadow models are trained in the same setup as the target model; thus, it is reasonable to expect the alignment in the behaviour of the target model and the shadow ones (up to the inherent randomness in the training process, e.g., random augmentations). However, in this setting, an attacker does not explicitly facilitate shadow models to capture the exact target model's behaviour, such as confidence scores, making the degree of alignment. Furthermore, as we show in Figure \ref{fig:method_distribution}, if the architectures of the target model and shadow models are not the same, the divergence between their outputs may be significant, resulting in a wrong estimation of the target model's output distribution and reducing the final attack performance.

To address these limitations, we propose leveraging knowledge distillation as an approach to train the shadow models in the context of MIAs. We emphasize that this is beneficial for an attacker since the knowledge distillation can yield a higher degree of similarity in the predictions between the target model and shadow models \cite{kim2021comparing}.  

\subsection{Likelihood Ratio Attack}


Likelihood Ratio Attack (LiRA, \cite{carlini2022membership}) is, according to Neyman–Pearson lemma \cite{neyman1933ix}, the state-of-the-art membership inference attack that leverages information only about the outputs of the target model. Likelihood Ratio Attack may be set up in two different setting, namely, in \emph{online} and \emph{offline} ones.

\subsubsection{Online LiRA} In this setting, an attacker first trains $N$ shadow models on random samples from the data distribution $\mathbb{D}$, so that for the given target point $(x^*, y^*)$ is included in the training sets of exactly half of these models (IN models), and is not presented in the training sets of the other half (OUT models). Then, an attacker computes the confidence scores of IN models and OUT models are calculated given sample $(x^*, y^*)$. These scores are computed as the logits of the model's predictions in the form

\begin{equation}
    \phi(p) = \log\left(\frac{p}{1-p}\right), \ \text{where } p = f_{\theta}(x)_y.
\end{equation}
Given a sufficiently large number of well-trained shadow models, the distribution of $\phi(p)$ is approximately Gaussian \cite{carlini2022membership}. Thus, after retrieving the scores for IN and OUT models, the attacker fits two Gaussian distributions, denoted as $\mathcal{N}(\mu_{\text{in}}, \sigma_{\text{in}}^2)$ and $\mathcal{N}(\mu_{\text{out}}, \sigma_{\text{out}}^2)$ to approximate the densities of $\phi(p \vert \text{in})$ and $\phi(p \vert \text{out})$, respectively. Then, the attacker queries the target  model $f^{t}_\theta$ with $(x^*, y^*)$ and computes the corresponding logit in the  way described above to obtain 

\begin{equation}
    \text{conf}_{\text{obs}} = \phi(p^{*}), \ \text{where} \  p^{*} = f^{t}_\theta(x^{*})_{y^*}.
\end{equation} 
Finally, using the likelihood ratio test between the two hypotheses, an attacker computes the score 
\begin{equation}
   s(x^*, y^*) = \frac{p(\text{conf}_{\text{obs}} \ | \ \mathcal{N}(\mu_{\text{in}}, \sigma_{\text{in}}^2))}{p(\text{conf}_{\text{obs}} \ | \ \mathcal{N}(\mu_{\text{out}}, \sigma_{\text{out}}^2))},
\end{equation}
that represents the confidence of LiRA to assign $(x^{*}, y^{*})$ to $D_t.$ Here, $p(\text{conf}_{\text{obs}} \ | \ \mathcal{N}(\mu, \sigma^2))$ is the probability density function of  $\text{conf}_{\text{obs}}$ under $\mathcal{N}(\mu, \sigma^2)$.

\subsubsection{Offline LiRA} For the offline threat model, the attacker only trains OUT shadow models, and the final score is calculated using a one-sided hypothesis test in the form 
\begin{equation}
\label{eq:offline_lira}
    s(x^{*}, y^{*}) = 1 - p(\text{conf}_{\text{obs}} \ | \ \mathcal{N}(\mu_{\text{out}}, \sigma_{\text{out}}^2)).    
\end{equation}
In this setting, an attacker does not calculate the distribution of IN scores $\mathcal{N}(\mu_{\text{in}}, \sigma_{\text{in}}^2)$. Although an attacker leverages less information to determine the membership status of the target point $(x^*, y^*)$, the setup avoids training new shadow models at the inference time for each newly received target point. Note that in both scenarios, the attacker do not leverage any information about the target model. 

\subsection{Guided Likelihood Ratio Attack (GLiRA)}

\begin{algorithm}[t]
 \begin{algorithmic}[1]
  \REQUIRE $\text{Dataset} \ D_\text{shadow}$, $\text{target model} \ f^t$, $\text{number of training steps} \ T$, $\text{learning rate} \ \eta$
  \STATE $f^s_{\theta_0} \gets \text{Initialize model by } \theta_0 \sim \Theta$
  \FOR{$k = 1,...,T$}
     \STATE Sample batch of data $B \sim D_\text{shadow}$
    \STATE $\mathcal{L}_{\text{KD}} \gets \frac{1}{\vert B \vert}  \sum_{(x,y) \in B} \mathcal{L}_{\text{KD}}(y,f^t, f^s_\theta)$
    \COMMENT {Compute the loss function in the form of Eq. \eqref{eq:knowledge_distillation_kl} or Eq. \eqref{eq:knowledge_distillation_mse}}
     \STATE $\theta_k \gets \theta_{k-1} - \eta \nabla_{\theta_{k-1}} \mathcal{L}_{\text{KD}}$
  \ENDFOR
  \vspace{0.5em}
    \RETURN $ \displaystyle f_{\theta_T}^s$
 \end{algorithmic}
 \caption{Distill}
 \label{alg:distill}
\end{algorithm}

\begin{algorithm}[t]
 \begin{algorithmic}[1]
  \REQUIRE $\text{Target model} \ f^t$, $\text{data point} \ (x^*, y^*)$, $\text{data distribution} \ \mathbb{D}$, $\text{number of training steps} \ T$, $\text{learning rate} \ \eta$
  
  \STATE $\text{confs}_{\text{out}} = \{\}$
  \FOR{\text{$1,...,N$}}
    \STATE $D_{\text{shadow}} \sim \mathbb{D}$
    \STATE $f_{\text{out}} \leftarrow 
\texttt{Distill} (D_{\text{shadow}}, f^t, T, \eta)$
    \STATE $\text{confs}_{\text{out}} \leftarrow \text{confs}_{\text{out}} \cup \{\phi(f_{\text{out}}(x^*)_{y^*})\}$
  \ENDFOR
  \STATE $\mu_{\text{out}} \gets \texttt{mean}(\text{confs}_{\text{out}})$
  \STATE $\sigma_{\text{out}}^2 \gets \texttt{var}(\text{confs}_{\text{out}})$
  \STATE $\text{conf}_{\text{obs}} = \phi(f(x^*)_{y^*})$
  \vspace{0.5em}
  \RETURN $\displaystyle { 1 - p(\text{conf}_{\text{obs}}\ \mid\ \mathcal{N}(\mu_{\text{out}}, \sigma^2_{\text{out}}))}$
 \end{algorithmic}
 \caption{Guided Likelihood Ratio Attack}
 \label{alg:glira}
\end{algorithm}


In this section, we describe the proposed approach to \emph{guided} training of shadow models. 
Instead of utilizing a default loss function to train shadow models, which does not explicitly force them to mimic the outputs of the target model, we incorporate knowledge distillation into their training. We do this by modelling the setting from Section \ref{subsec:kd}, assuming that the target model is the ``teacher'' and each shadow model is a ``student''. In this setting, we can use KD framework to train each shadow network to mimic the outputs of the target network.


In our setting, we consider the \textbf{offline} scenario of LiRA.
This decision is motivated by two factors. First of all, we do not need to train additional (IN) shadow models for each newly coming target point, what makes it possible to set up the membership inference attack pipeline only once. Secondly, within the KD framework, the similarity in behaviour between the target model and shadow models is of higher priority than the prediction of the latter on specific data samples. Thus, the shadow models may not be able to capture the subtle differences in outputs between training members and non-members, which is critical for informative IN models. At the same time, the goal of  OUT models is to mimic the behaviour of the target network on the unseen data, which can be achieved by using the KD framework.

In our framework, we train $N$ OUT shadow models of the given architecture $f^s$  (see the procedure  $\texttt{Distill}$, Algorithm \ref{alg:distill}). During training, we minimize the $\mathcal{L_{\text{KD}}}$ loss in the form from Eq. \eqref{eq:knowledge_distillation_kl} or Eq. \eqref{eq:knowledge_distillation_mse} on random samples from the data distribution $\mathcal{D}$. Then, given a target point $(x^*, y^*)$, we calculate confidence scores of OUT models and fit a Gaussian $\mathcal{N}(\mu_{\text{out}}, \sigma_{\text{out}}^2)$ to estimate their distribution. Finally, by querying the target model, we compute the score in the form from Eq. \eqref{eq:offline_lira} as the confidence of our approach to assigning the sample  $(x^*, y^*)$ to the training set of the target model. 

The pseudo-code describing the proposed approach is presented in Algorithm \ref{alg:glira}. The proposed pipeline for shadow models training is presented in Figure \ref{fig:enter-label}.


\section{Experiments}

\begin{figure*}[t]
    \centering
    \includegraphics[width=0.45\textwidth]{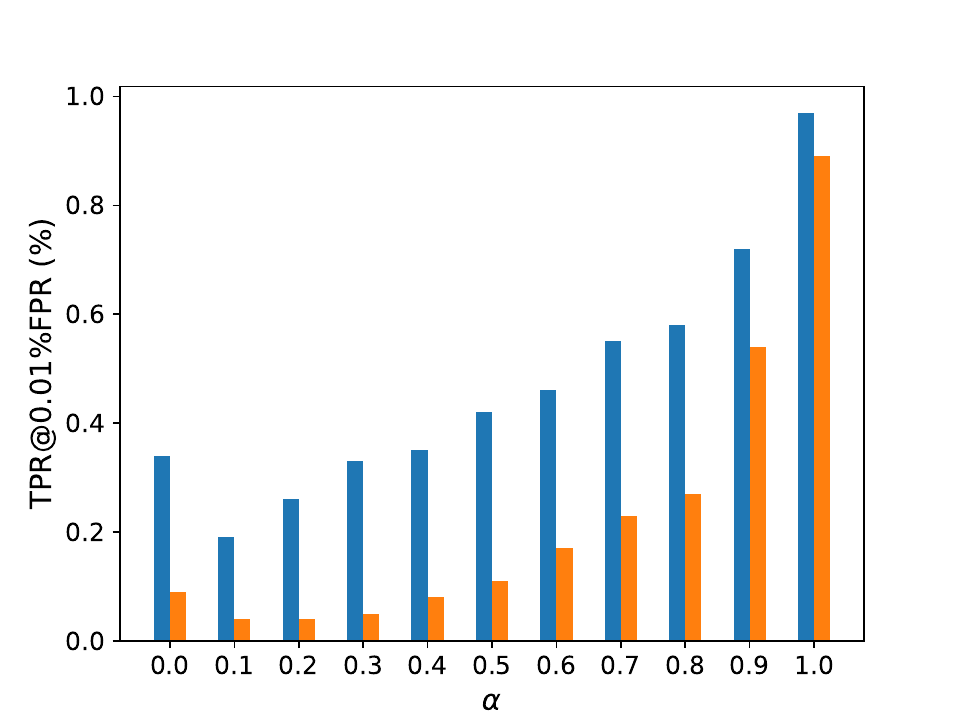}
    \includegraphics[width=0.45\textwidth]{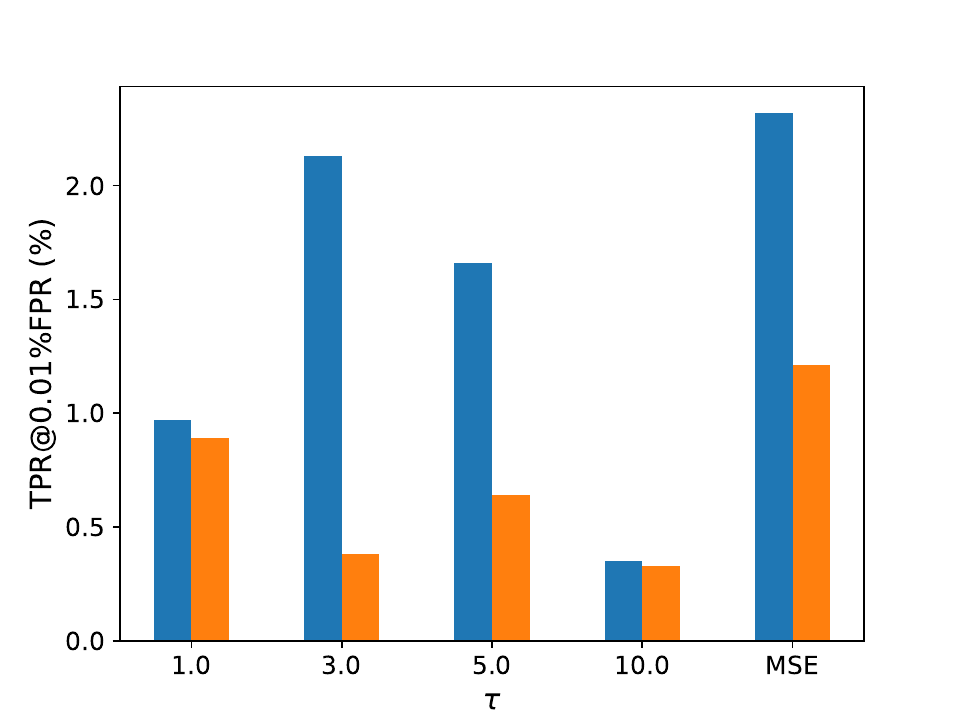}
    \caption{The effect of the balancing factor $\alpha$ and the temperature parameter $\tau$ from Eq. \eqref{eq:knowledge_distillation_kl} on the success rate of the proposed attack methods. We present results on a fixed low FPR rate of $0.01 \%$ and consider two experimental setups. \textbf{Blue}: the architecture of target and shadow models is the same (namely, MobileNet-V2). \textbf{Orange}: the architecture of the target model is MobileNet-V2; the architecture of shadow models is ResNet34.}
    \label{fig:kl_weight}
\end{figure*}

In this section, we describe the evaluation of our guided likelihood ratio attack.

\subsection{Experimental Setting}

\subsubsection{Datasets and Training} 
\label{subsec:training}



We utilize several datasets traditionally used to evaluate the effectiveness of the membership inference attack methods: CIFAR10, CIFAR100 \cite{krizhevsky2009learning}, CINIC10 \cite{darlow2018cinic10}. When training the models on CINIC10, we remove the CIFAR10 part from it. To ensure that the training sets of the target model and shadow models do not intersect, we split the initial dataset into three non-intersecting parts. Given the training dataset $D$, for CIFAR10 and CIFAR100 datasets, the target model is trained on the random subset $D_t$ of $20000$ samples; each shadow model is trained on a random subset $D_\text{shadow}$ of size $20000$ samples from the dataset $D \setminus D_t$; remaining $10000$ from the test dataset $D_\text{test}$ are marked as non-members and used in the evaluation of the method (note that to balance members and non-members during the evaluation, we sample random subset of size $10000$ from $D_t$ and mark them as members). Similarly, for CINIC10 dataset, the training dataset $D_t$ of the target model is of size  $50000$, and each shadow model is trained on a random subset of size $50000$ sampled from the remaining dataset; the remaining  $50000$ samples from the test dataset are marked as non-members, and its random subset of $20000$ together with a random subset from $D_t$ of the same size are used in the evaluation of the method.

All models were trained for $100$ epochs to achieve reasonably high classification accuracy (namely, 87.0-91.0\% for CIFAR10, 57.27-66.7\% for CIFAR100, and 78.0-82.0\% for CINIC10 depending on the architecture). We used SGD optimizer with learning rate of $0.1$, weight decay of $0.5 \times 10^{-3}$ and momentum of $0.9$. For each experiment, we train $128$ shadow models. Following the evaluation protocol from the other works, we query the target model on multiple points obtained by applying standard data augmentations to the target point. The number of queries is set to $10$.

\begin{figure*}[ht!]
    \centering
    \includegraphics[width=0.32\textwidth]{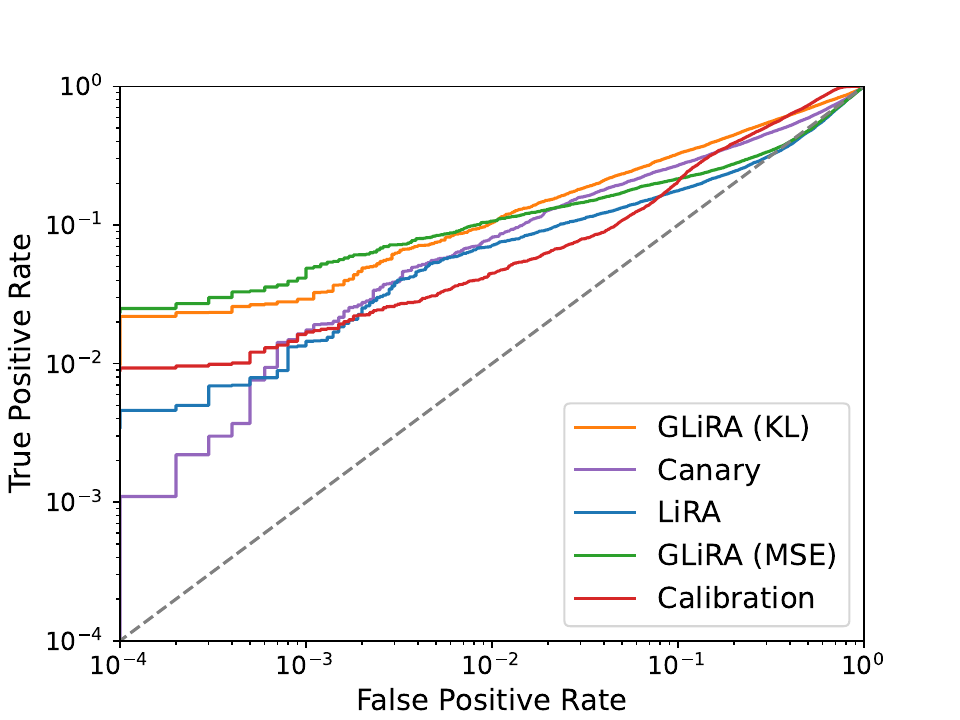}
    \includegraphics[width=0.32\textwidth]{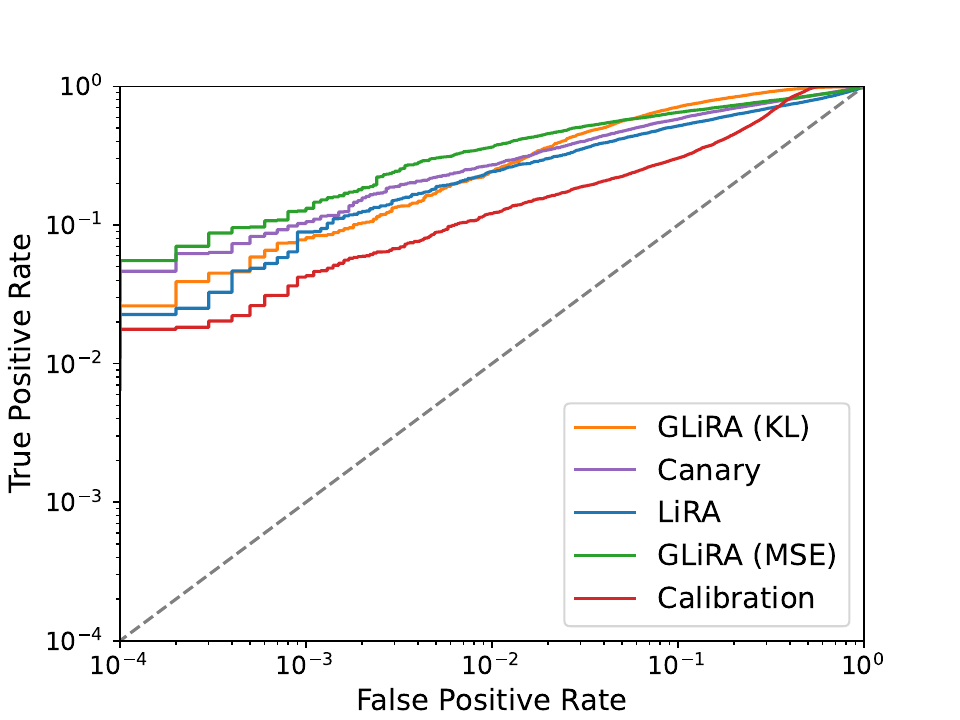}
    \includegraphics[width=0.32\textwidth]{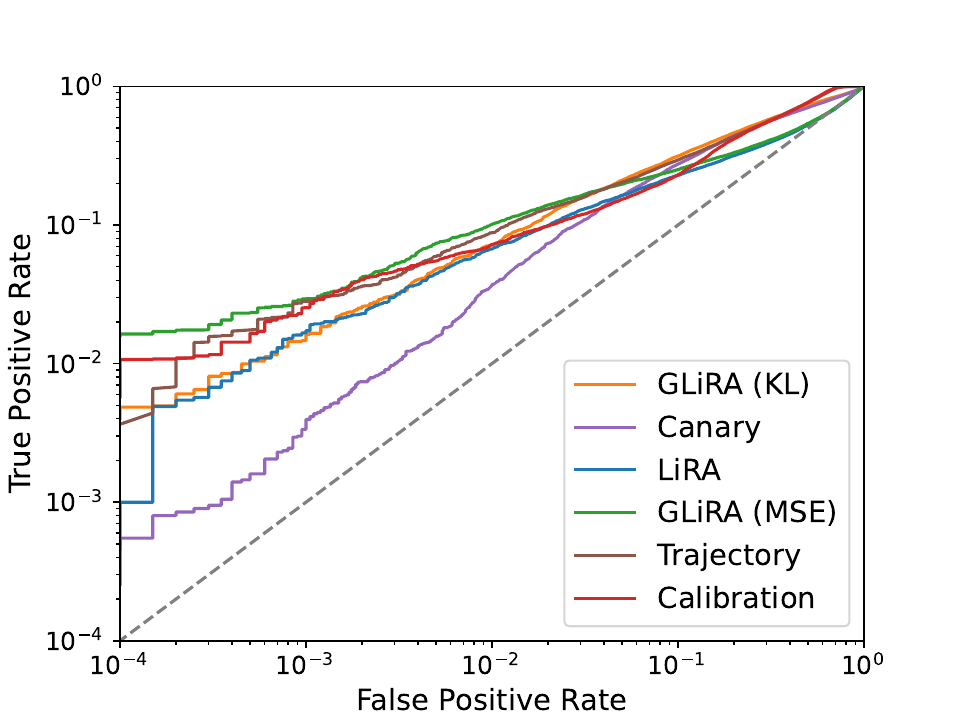}

    \includegraphics[width=0.32\textwidth]{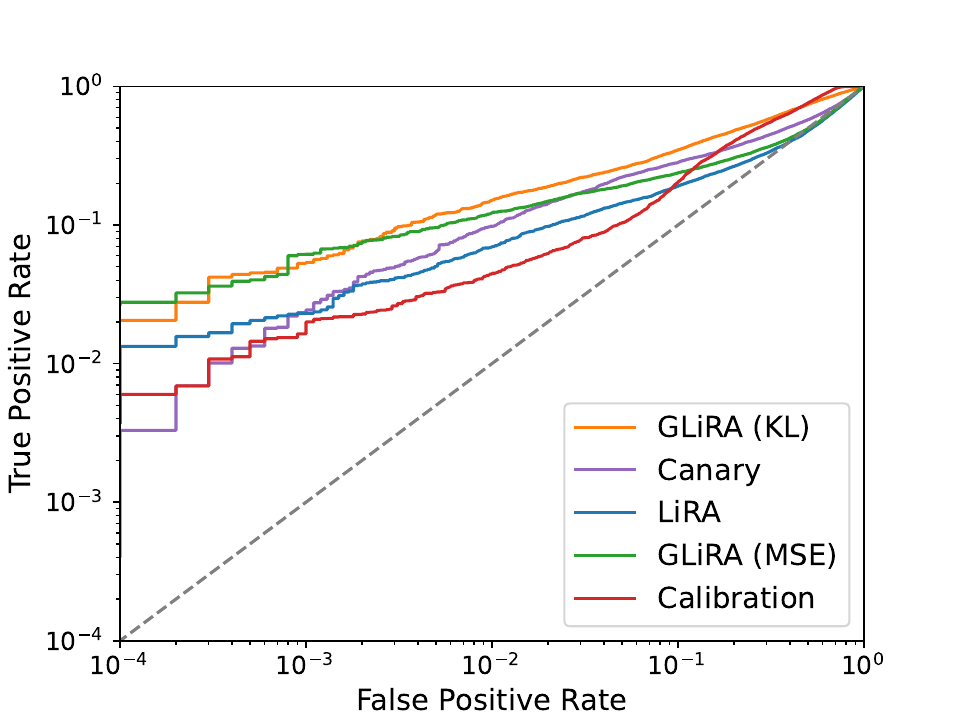}
    \includegraphics[width=0.32\textwidth]{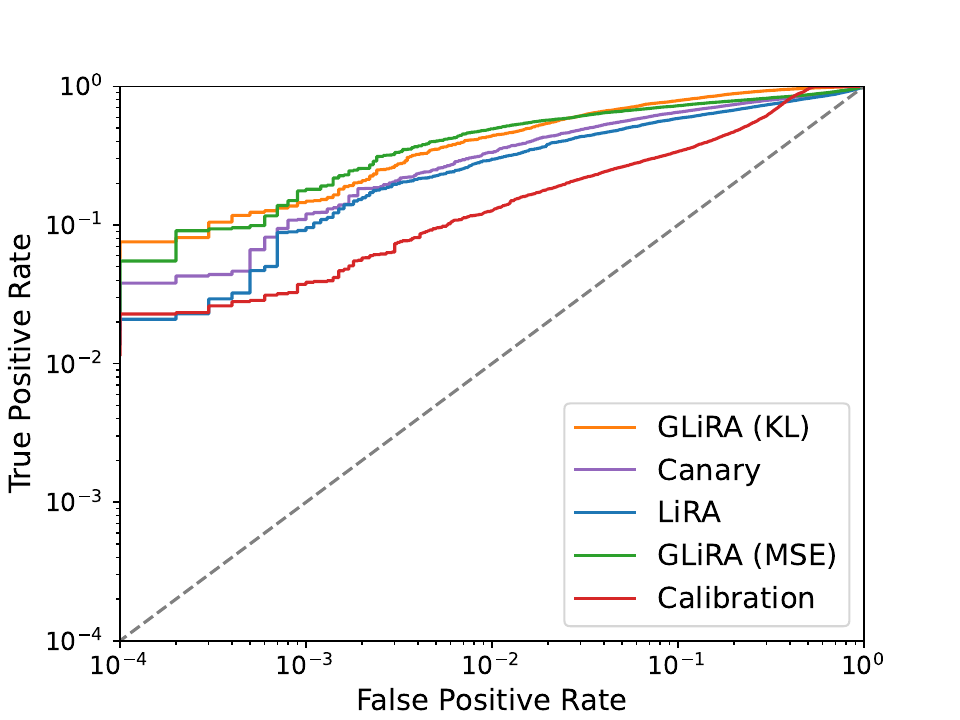}
    \includegraphics[width=0.32\textwidth]{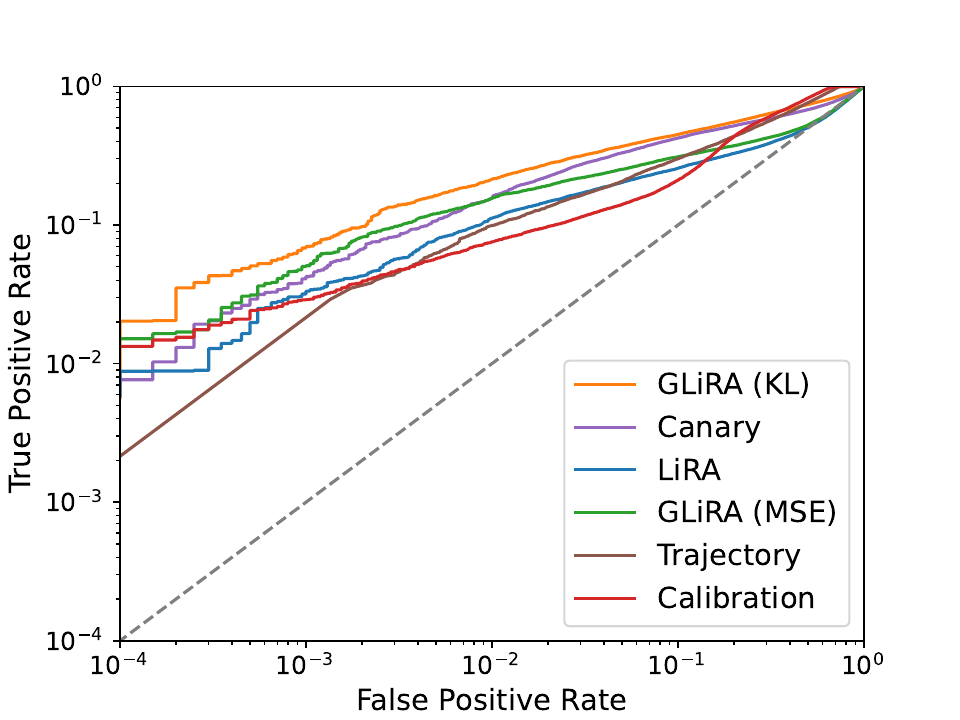}
    
    \includegraphics[width=0.32\textwidth]{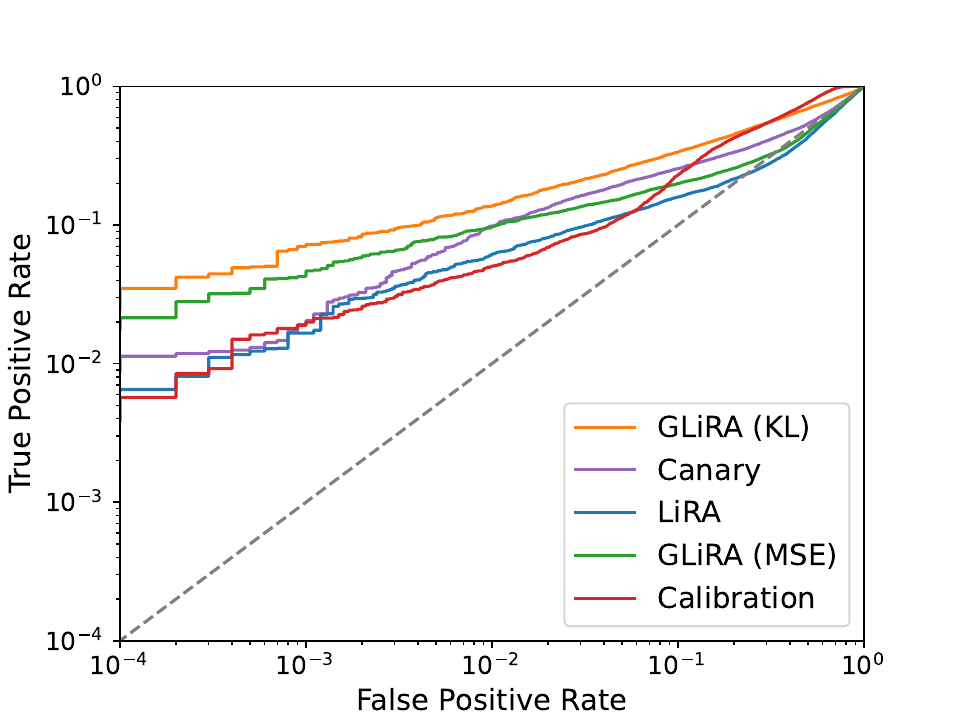}
    \includegraphics[width=0.32\textwidth]{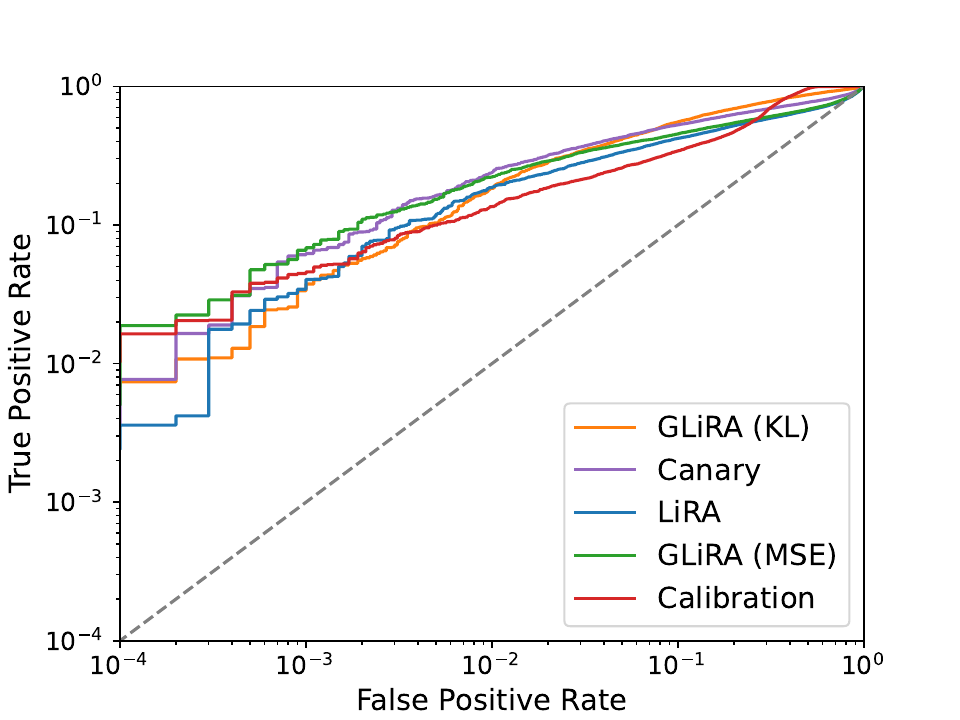}
    \includegraphics[width=0.32\textwidth]{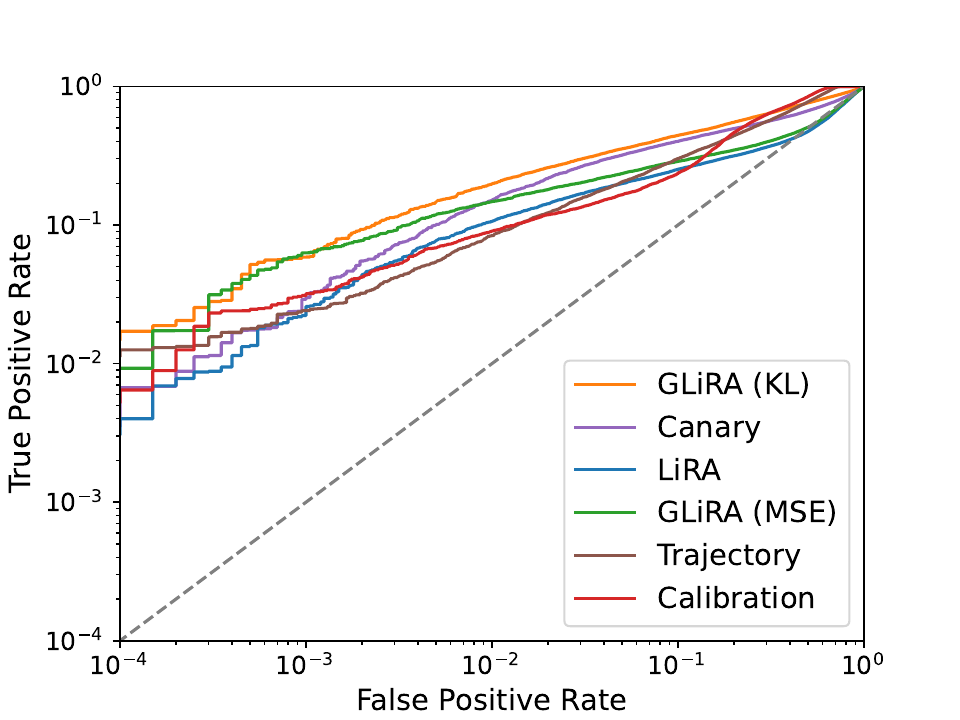}

    \subfloat[CIFAR10]{\includegraphics[width=0.32\linewidth]{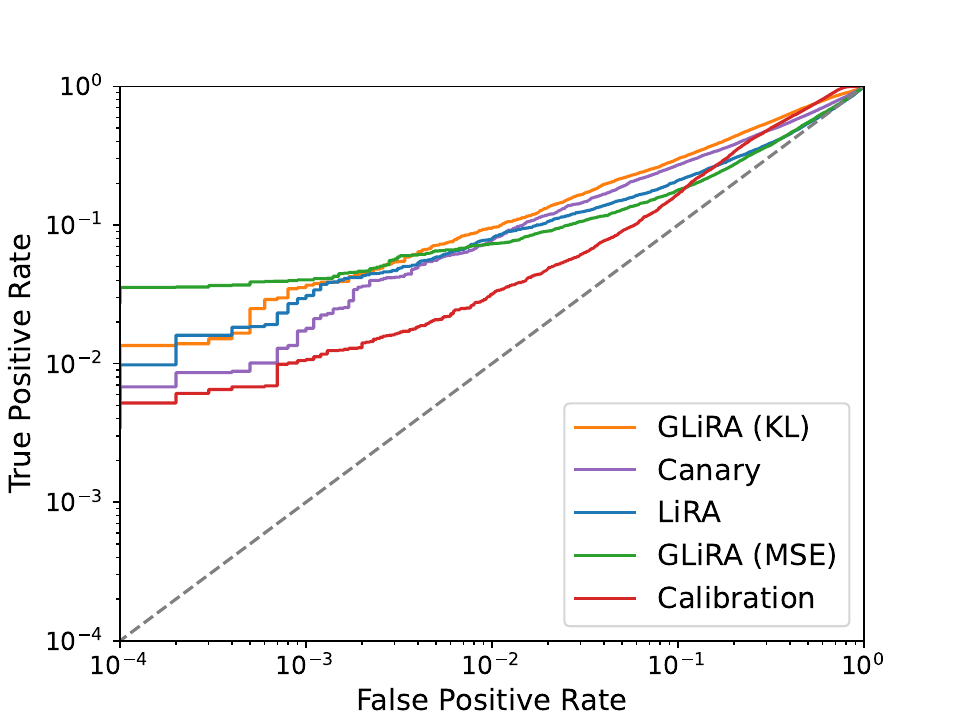}}\hspace{0.1mm}
    \subfloat[CIFAR100]{\includegraphics[width=0.32\linewidth]{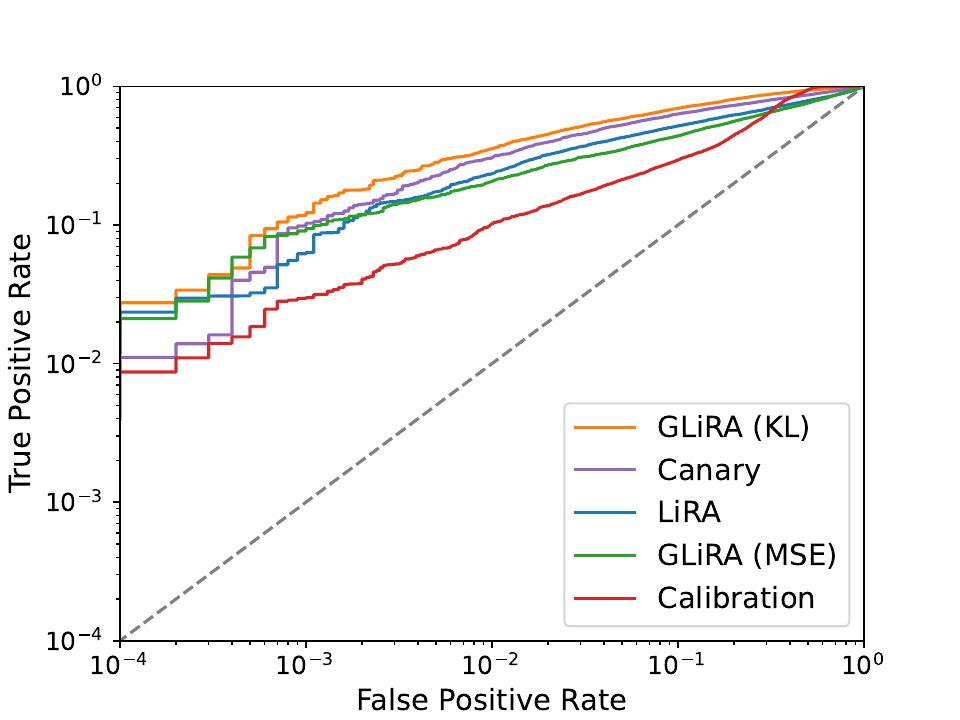}}\hspace{0.1mm}
    \subfloat[CINIC10]{\includegraphics[width=0.32\linewidth]{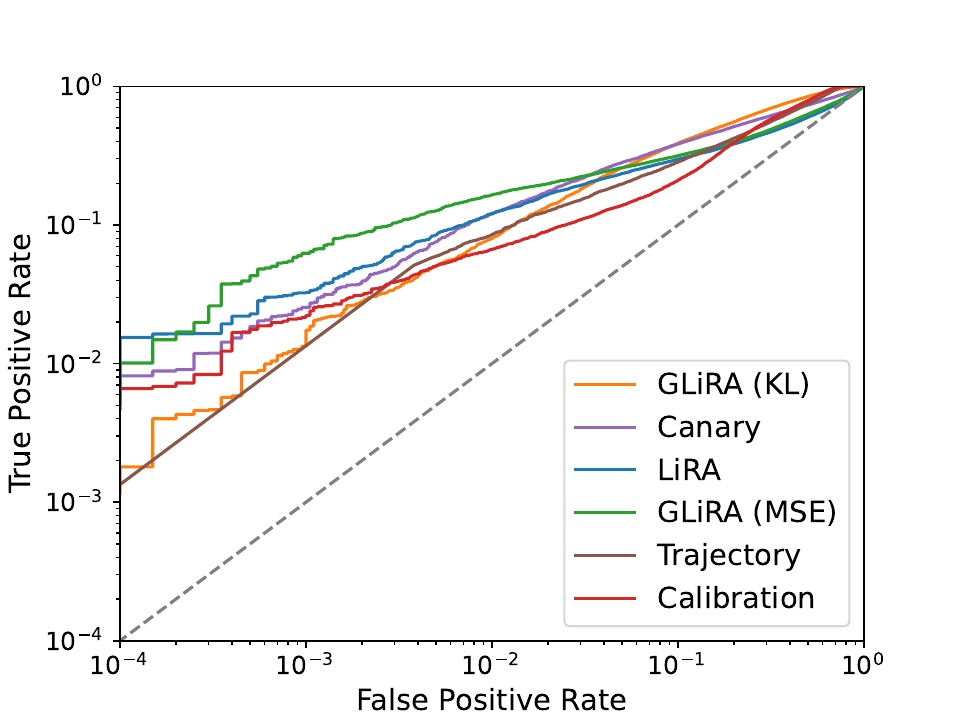}}\hspace{0.1mm}
    
    \caption{The quantitative results of experiments. We compare the performance of different attack methods in the setting when the adversary is aware of the target model architecture and uses it to train shadow models. Results are presented for  three different datasets and four model architectures (from top to bottom: MobileNet-V2, ResNet-34, VGG16, WideResNet28-10).}
    \label{fig:main_same}
\end{figure*}

\begin{figure*}[ht!]
    \centering
    \includegraphics[width=0.32\textwidth]{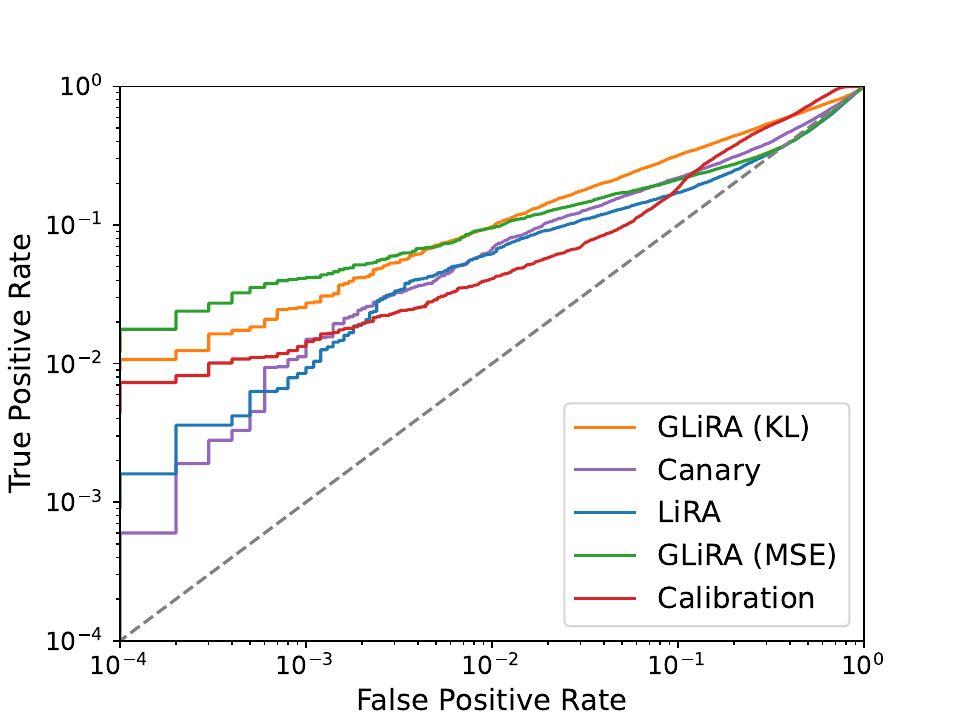}
    \includegraphics[width=0.32\textwidth]{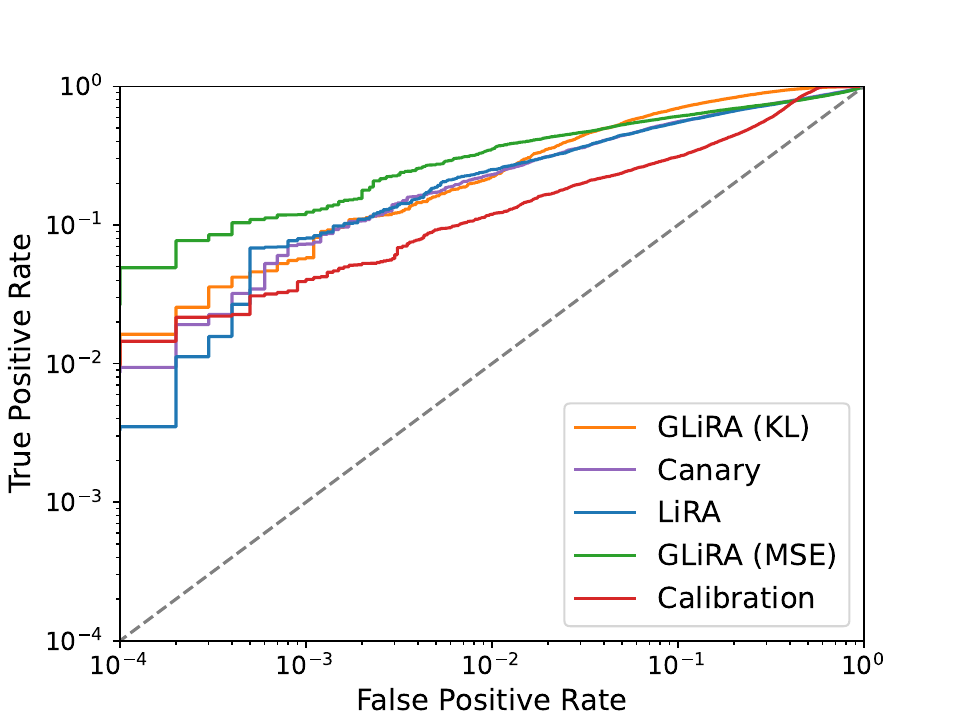}
    \includegraphics[width=0.32\textwidth]{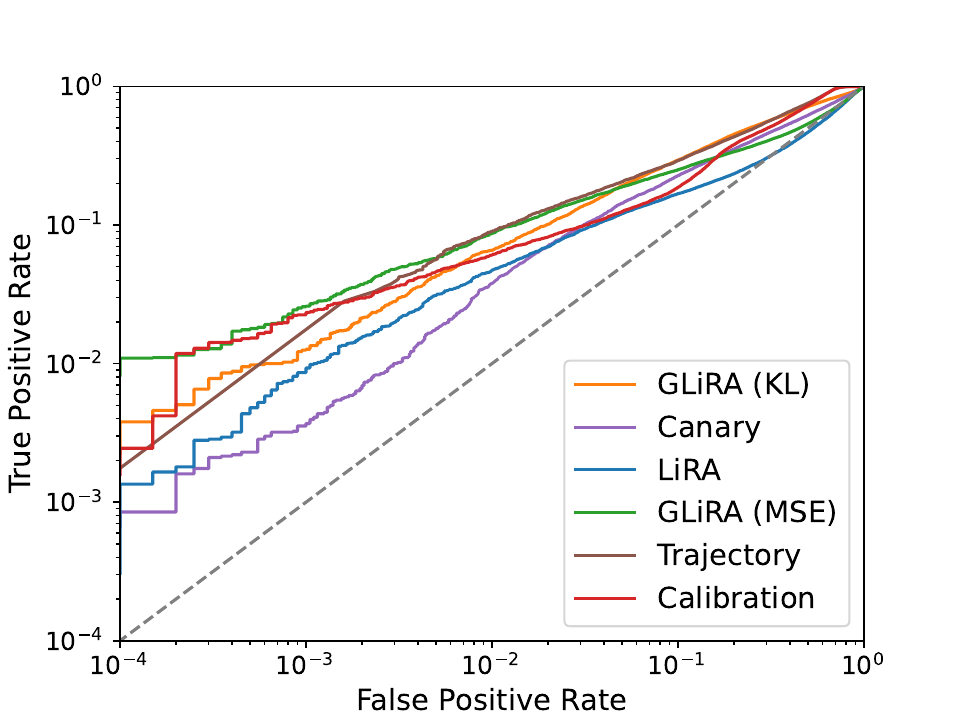}

    \includegraphics[width=0.32\textwidth]{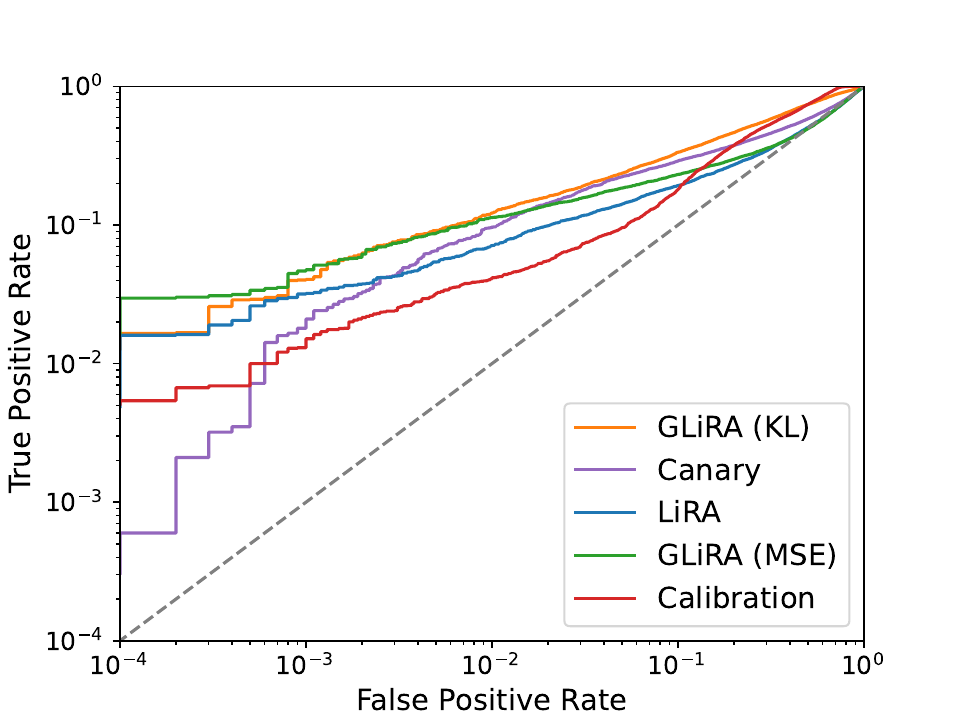}
    \includegraphics[width=0.32\textwidth]{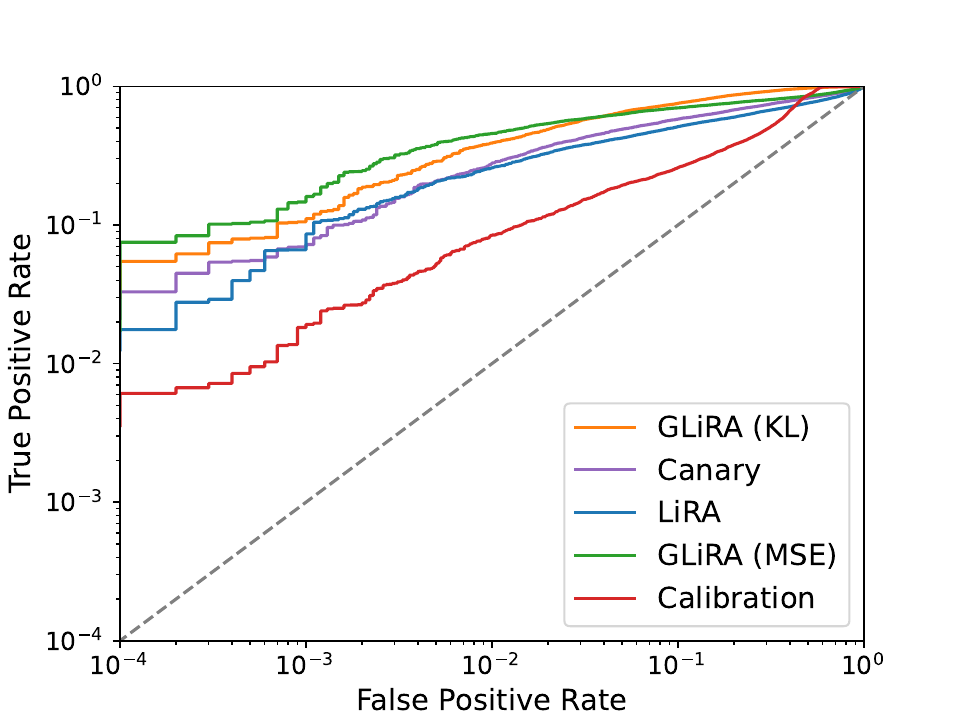}
    \includegraphics[width=0.32\textwidth]{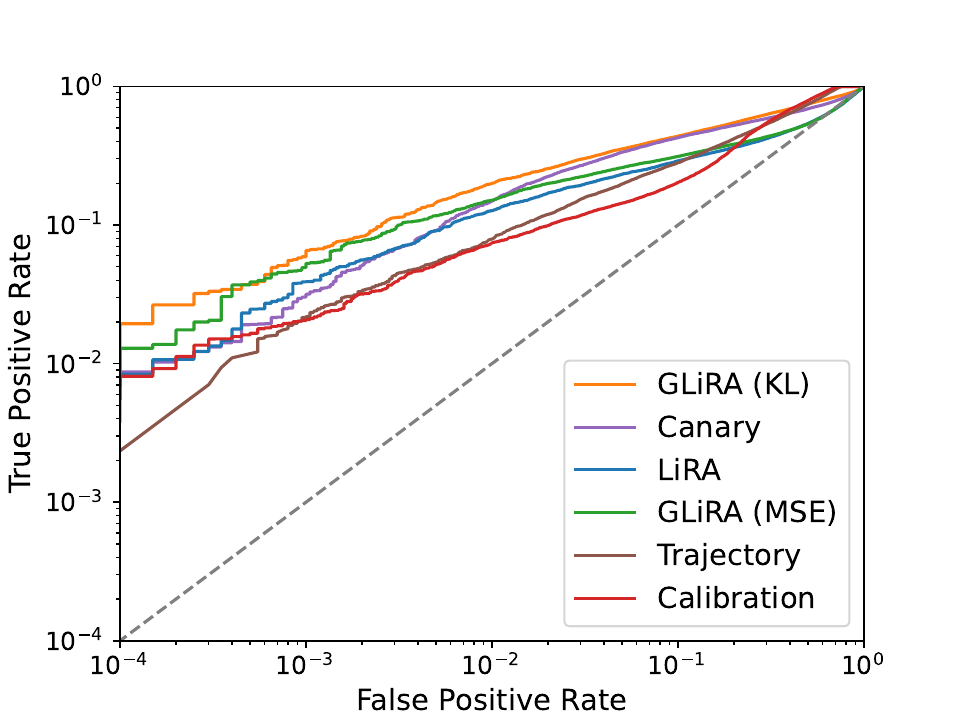}
    
    \includegraphics[width=0.32\textwidth]{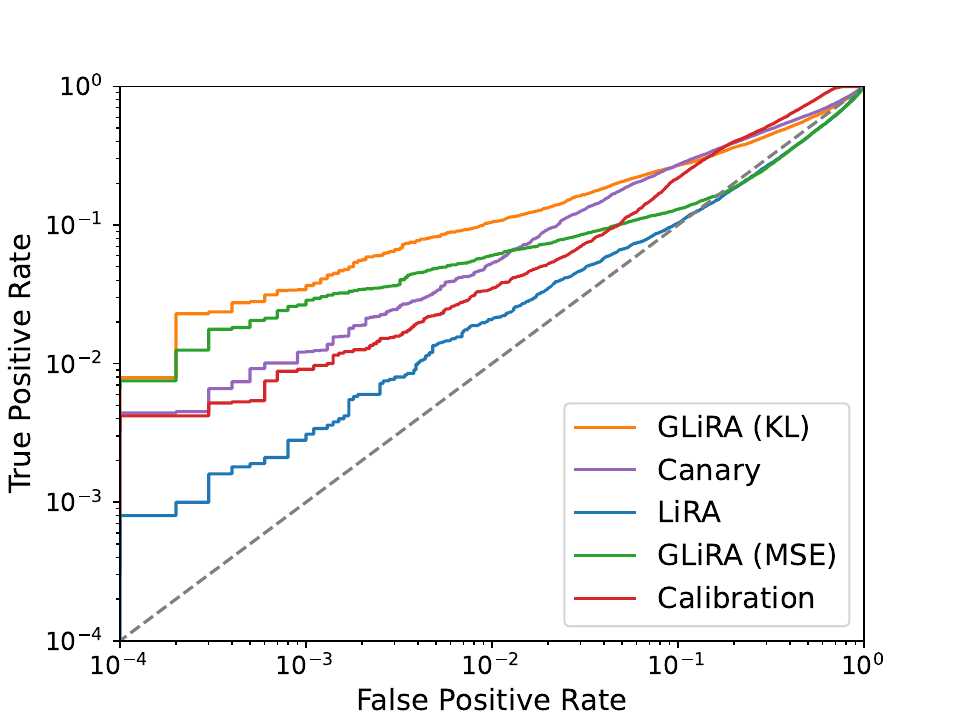}
    \includegraphics[width=0.32\textwidth]{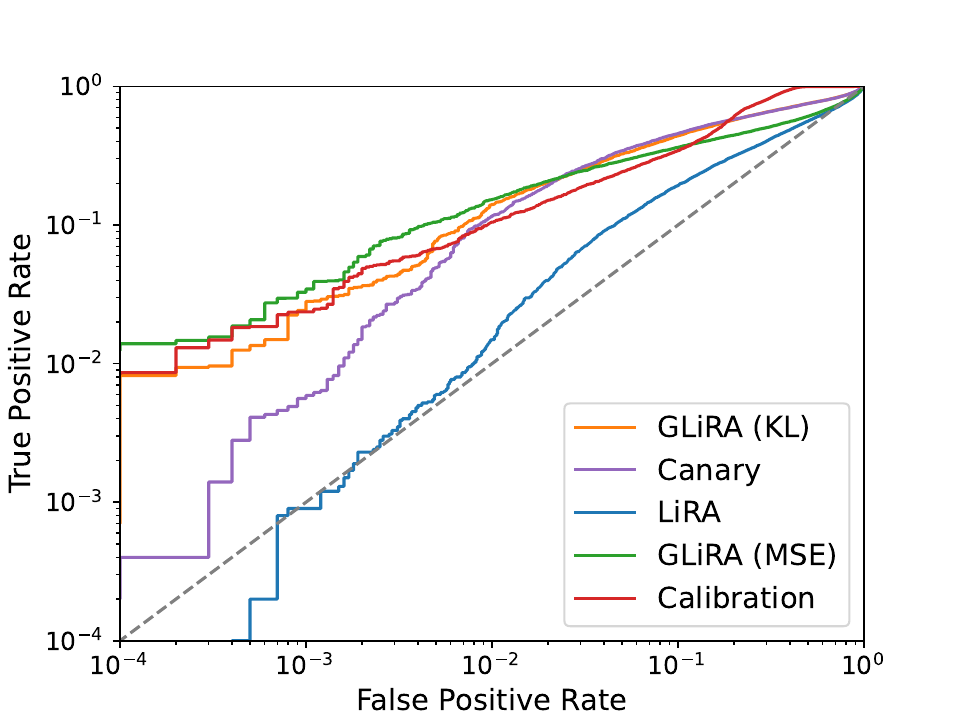}
    \includegraphics[width=0.32\textwidth]{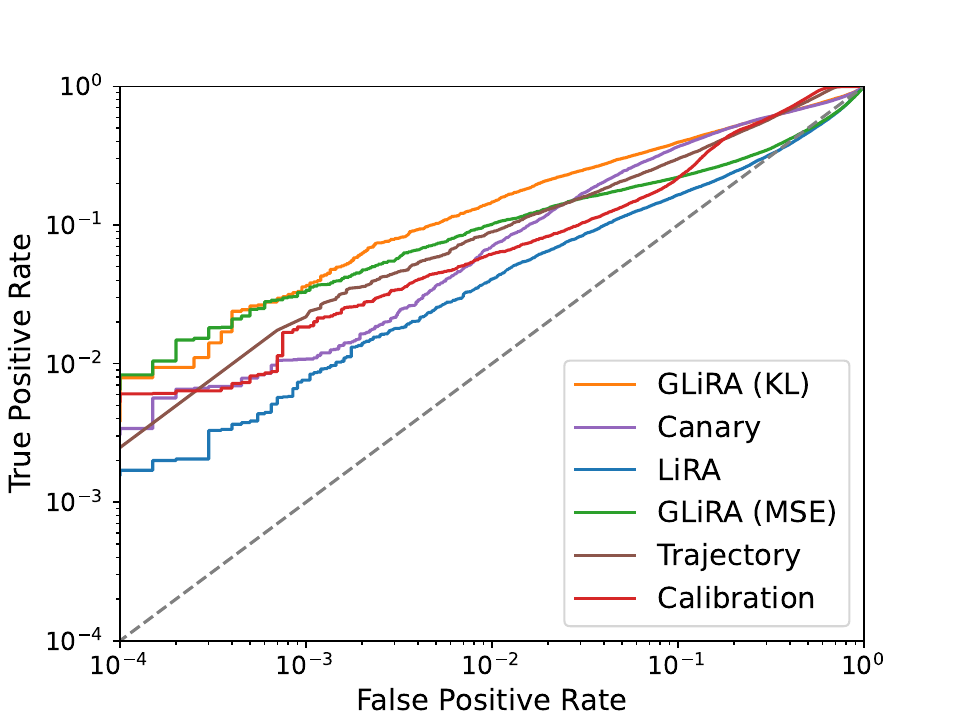}

    \subfloat[CIFAR10]{\includegraphics[width=0.32\linewidth]{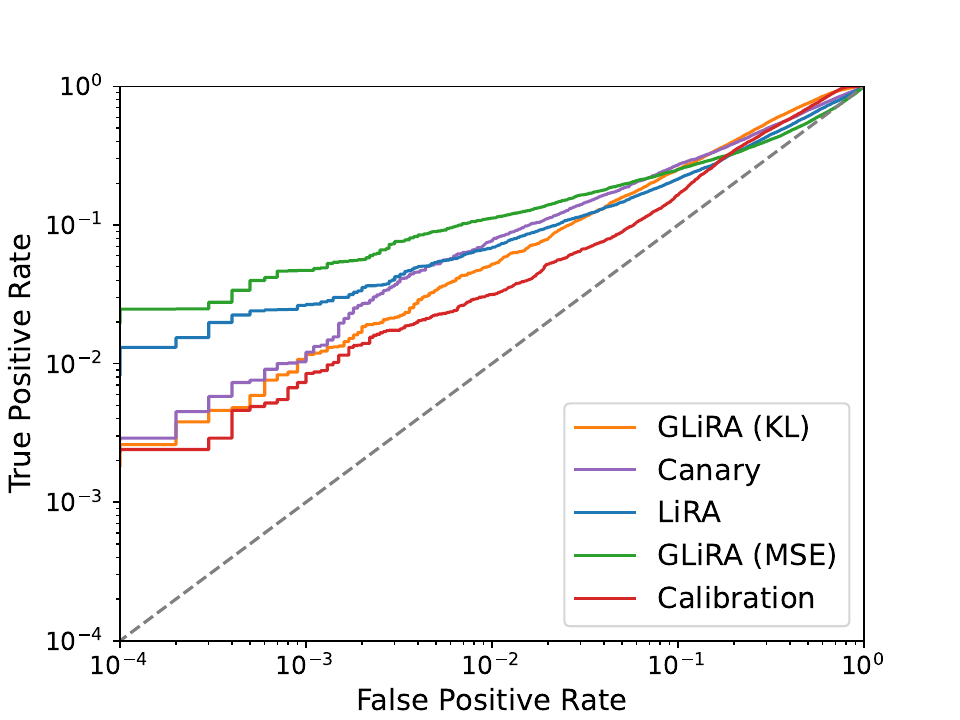}}\hspace{0.1mm}
    \subfloat[CIFAR100]{\includegraphics[width=0.32\linewidth]{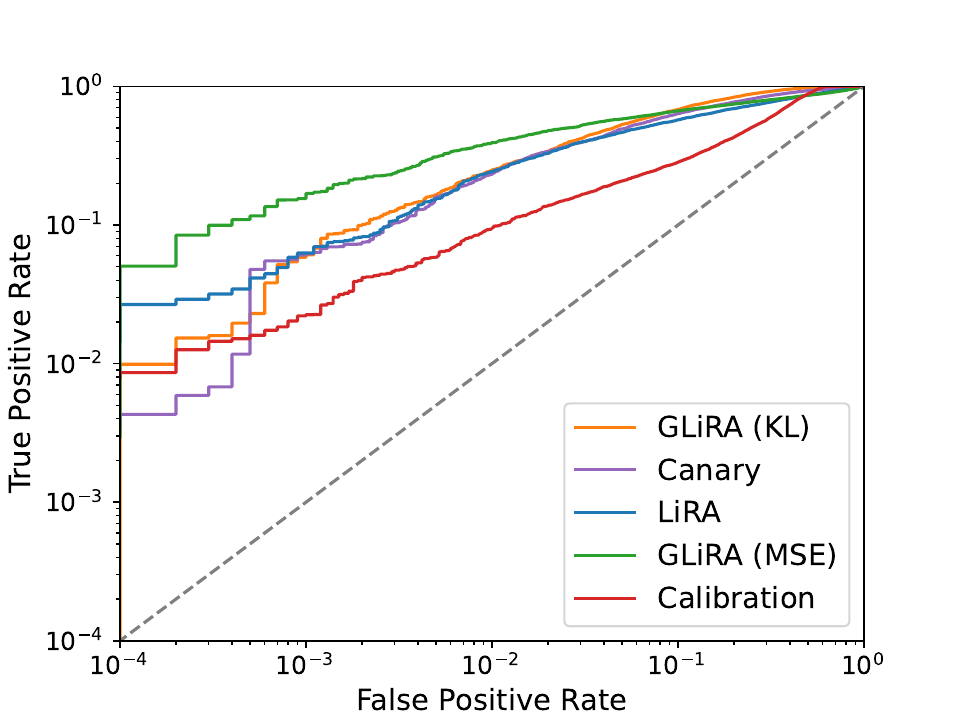}}\hspace{0.1mm}
    \subfloat[CINIC10]{\includegraphics[width=0.32\linewidth]{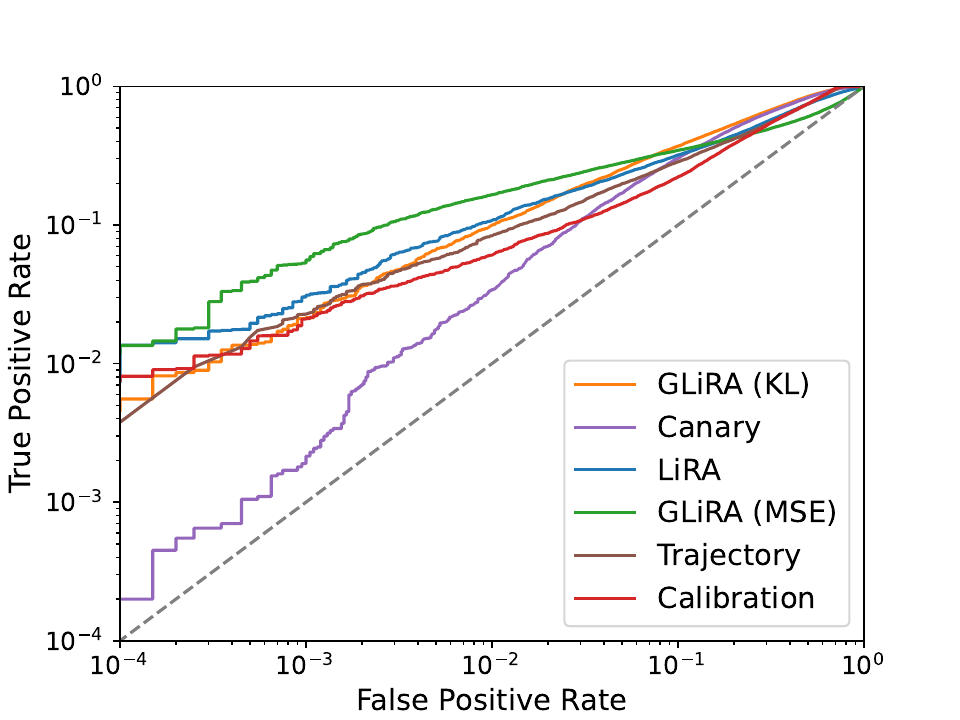}}\hspace{0.1mm}
    
    \caption{The quantitative results of experiments. We compare the performance of different attack methods in the setting when the adversary is unaware of the target model architecture and, hence, can not use it to train shadow models. Results are presented for  three different datasets and four model architectures  (from top to bottom: Target MobileNet-V2, Shadow ResNet-34; Target ResNet-34, Shadow VGG16; Target VGG16, Shadow WideResNet28-10; Target WideResNet28-10, Shadow MobileNet-V2).}
    \label{fig:main_diff}
\end{figure*}




\begin{table*}[ht!]
\centering
\caption{Performance of different membership inference attack methods. Target model's architecture is ResNet-34, shadow models' architecture is ResNet-34.}
\label{table:attack_res34_same}
\begin{tabular}{lcccccccccccc}
\toprule
& \multicolumn{3}{c}{TPR at 0.01\% FPR}& \multicolumn{3}{c}{TPR at 0.1\% FPR}&  \multicolumn{3}{c}{TPR at 1\% FPR}& \multicolumn{3}{c}{AUC} \\

\cmidrule(l{5pt}r{5pt}){2-4}\cmidrule(l{5pt}r{5pt}){5-7}\cmidrule(l{5pt}r{0pt}){8-10}\cmidrule(l{5pt}r{0pt}){11-13}
\multirow{1}{*}{}  & CF10& CF100& CINIC10& CF10& CF100& CINIC10& CF10& CF100& CINIC10& CF10& CF100& CINIC10\\
\midrule
LiRA~\cite{carlini2022membership}&0.37\%&1.36\%&0.57\%&2.30\%&9.13\%&3.18\%&6.93\%&29.80\%&11.19\%&0.510&0.787&0.540\\
Calibration~\cite{watson2021importance}&0.37\%&1.14\%&1.33\%&1.64\%&3.73\%&2.88\%&4.43\%&12.62\%&7.55\%&0.690&0.779&\textbf{0.724}\\
Canary~\cite{DBLP:conf/iclr/WenBKBGGG23}&0.04\%&1.74\%&0.70\%&2.33\%&10.96\%&4.09\%&9.81\%&33.45\%&15.58\%&0.580&0.827&0.669\\
Trajectory~\cite{LZBZ22}&-&-&0.00\%&-&-&0.00\%&-&-&9.92\%&-&-&0.710\\
\midrule
GLiRA (KL)&\textbf{2.00\%}&\textbf{5.73\%}&0.56\%&\textbf{5.29\%}&\textbf{14.49\%}&\textbf{6.83\%}&\textbf{15.06\%}&\textbf{43.83\%}&\textbf{21.41\%}&\textbf{0.694}&\textbf{0.925}&0.711\\
GLiRA (MSE)&\textbf{2.62\%}&\textbf{2.23}\%&\textbf{1.50\%}&\textbf{6.14\%}&\textbf{17.62\%}&\textbf{5.02}\%&\textbf{12.19}\%&\textbf{49.41\%}&15.55\%&0.534&\textbf{0.854}&0.566\\
\bottomrule
\end{tabular}
\end{table*}

\begin{table*}[ht!]
\centering
\caption{
Performance of different membership inference attack methods. Target model's architecture is ResNet-34, shadow models' architecture is VGG16.}
\label{table:attack_res34_diff}
\begin{tabular}{lcccccccccccc}
\toprule
 & \multicolumn{3}{c}{TPR at 0.01\% FPR}& \multicolumn{3}{c}{TPR at 0.1\% FPR}&  \multicolumn{3}{c}{TPR at 1\% FPR}& \multicolumn{3}{c}{AUC} \\
\cmidrule(l{5pt}r{5pt}){2-4}\cmidrule(l{5pt}r{5pt}){5-7}\cmidrule(l{5pt}r{0pt}){8-10}\cmidrule(l{5pt}r{0pt}){11-13}
 & CF10& CF100& CINIC10& CF10& CF100& CINIC10& CF10& CF100& CINIC10& CF10& CF100& CINIC10\\
\midrule
LiRA~\cite{carlini2022membership}&0.48\%&1.23\%&0.69\%&3.18\%&6.62\%&3.89\%&7.09\%&25.80\%&12.67\%&0.522&0.733&0.562\\
Calibration~\cite{watson2021importance}&0.50\%&0.35\%&0.55\%&1.30\%&1.82\%&2.04\%&4.09\%&8.42\%&7.41\%&0.677&0.720&0.700\\
Canary~\cite{DBLP:conf/iclr/WenBKBGGG23}&0.03\%&2.08\%&0.68\%&1.80\%&6.95\%&3.02\%&9.63\%&27.76\%&14.86\%&0.587&0.787&0.673\\
Trajectory~\cite{LZBZ22}&-&-&0.00\%&-&-&2.15\%&-&-&7.91\%&-&-&0.697\\
\midrule
GLiRA (KL)&1.25\%&\textbf{2.93\%}&0.38\%&4.01\%&10.54\%&\textbf{5.91\%}&\textbf{12.18\%}&39.00\%&\textbf{19.90}\%&\textbf{0.689}&\textbf{0.916}&\textbf{0.710}\\
GLiRA (MSE)&\textbf{1.63\%}&1.70\%&\textbf{0.89\%}&\textbf{4.66\%}&\textbf{14.65\%}&4.93\%&11.33\%&\textbf{45.67\%}&15.10\%&0.531&0.837&0.571\\
\bottomrule
\end{tabular}
\end{table*}

\subsubsection{Evaluation Protocol}

In this work, we consider the following evaluation metrics:
\begin{itemize}
    \item \textit{AUC}. This is the metric for evaluation of classification tasks, which has also been widely applied for MIA (\cite{shokri2017membership, nasr2019comprehensive, watson2021importance, ye2022enhanced}). In practice, AUC is not very informative when measuring the success rate of a membership inference attack \cite{carlini2022membership}. It is reported to ensure the completeness of the comparison with the other works.
    \item \textit{TPR at low FPR}. Following the work \cite{carlini2021extracting}, we report the values of the true positive rate (TPR) at the fixed low value of the false positive rate (FPR). 
\end{itemize}

\subsubsection{Concurrent Works}

We evaluate our approach against the following methods.
\begin{itemize}
    \item \textbf{Likelihood Ratio Attack} \cite{carlini2022membership}. We build our work upon the offline scenario of LiRA, which uses a parametric approach to estimate the distribution of the target model's outputs for a given sample. The membership status of the data point is determined via a likelihood ratio test. 
    \item \textbf{Difficulty Calibration in Membership Inference Attacks} \cite{watson2021importance}. In this work, the loss for the sample of interest is calibrated by the average loss of shadow models to obtain a membership score. All the shadow models are trained without this sample, representing the offline attack setup.
    \item \textbf{Canary in a Coalmine} \cite{DBLP:conf/iclr/WenBKBGGG23}. In this work, it was proposed to improve the LiRA attack by crafting an \textit{adversarial query} out of each given example such that it separates the distributions of IN and OUT models as much as possible. Like in LiRA, both online and offline setups are described, and we compare our approach with the latter one.
    \item \textbf{Membership Inference Attacks by Exploiting Loss Trajectory} \cite{LZBZ22}. In this work, the loss values of the shadow models from intermediate epochs are exploited together with the loss from the given target model to decide the membership status of a given example. Knowledge distillation is used to represent the intermediate states of the target model to apply the attack in a black-box setting.
\end{itemize}

\subsection{Results of Experiments}


\subsubsection{Knowledge Distillation for MIA}
\label{subsec:experiment_kl}

To maximize the success rate of the proposed attack method, we perform hyperparameter tuning. Namely, we tune the parameters $\alpha$ and $\tau$ from Eq. \eqref{eq:knowledge_distillation_kl}. Parameter $\alpha$ was sampled from the interval $[0, 1]$ with step size of $0.1$. We then fix the best value of $\alpha$ and vary $\tau$, choosing from the set $\{1.0, 3.0, 10.0\}$. Following \cite{kim2021comparing}, we experiment with replacing $\mathcal{L}_{\text{KL}}$ with $\mathcal{L}_{\text{MSE}}$ in Eq. \eqref{eq:knowledge_distillation_kl}, thereby obtaining a loss as in Eq. \eqref{eq:knowledge_distillation_mse}. To take into account the degree of knowledge a potential adversary has about the target model (namely, the knowledge of the architecture) and their possibility to use the shadow models of the corresponding architecture,  we perform two separate experiments on the CIFAR10 dataset. For the target architecture, we utilize MobileNet-V2 \cite{sandler2019mobilenetv2} in both cases; for the second experiment, the shadow models architecture is ResNet-34 \cite{he2015deep}. We present the results in Figure \ref{fig:kl_weight}.

\subsubsection{Attack Evaluation}
\label{subsec:experiment_all}

The evaluation of our attack and its comparison to other methods is conducted on various architectures and datasets.

To provide a thorough analysis of our attack capabilities, we employ four well-known architectures: MobileNet-V2 \cite{sandler2019mobilenetv2}, ResNet-34 \cite{he2015deep}, VGG16 \cite{simonyan2015deep} and WideResNet28-10 \cite{zagoruyko2016wide}. In Figure \ref{fig:main_same}, we present the results for the standard setting, where the adversary is aware of the architecture of the target model and can train the shadow models of the same architecture. In Table \ref{table:attack_res34_same}, we provide an in-depth analysis of the results for ResNet-34 architecture; we report true-positive rates at $0.01\%, \ 0.1\%, \ 1.0\%$ false-positive rates, as well as the AUC score. In Figure \ref{fig:main_diff}, we evaluate our attack in the setting where the adversary does not know the architecture of the target model (here, we train the shadow models of the different architecture). The results are reported in Table \ref{table:attack_res34_diff}. Note that in the notations, CF10 and CF100 correspond to CIFAR10 and CIFAR100, respectively.  Also note that in \cite{LZBZ22}, the authors used a very small dataset to train target models, while keeping the most samples for the distillation task. This setup is infeasible to reproduce for small datasets (CIFAR10 and CIFAR100) while keeping the target model dataset size as large as we do ($20000$ samples). To this end, we report the performance of this approach only on CINIC10 dataset (where the size of the dataset for the distillation is equal to $70000$ samples).

\section{Discussion}

\subsection{Choosing the Best Configuration}

We empirically observe that increasing the balancing factor $\alpha$ boosts the TPR at low FPR regardless of the potential adversary's knowledge of the target model's architecture. This means that optimising the benign distillation loss is more beneficial for the success of our attack. Therefore, we fix $\alpha = 1.0$.

In \cite{kim2021comparing}, the authors have shown that increasing the temperature parameter $\tau$ can benefit the logit matching between student and teacher networks. Similarly, we tested if increasing $\tau$ while keeping $\alpha = 1.0$ can further improve the success rate of our attack. For $\tau = 3.0$, the TPR at $0.01\%$ FPR does improve for the experiment when the architectures of the target model and shadow ones are the same (experiment A); however, this change did not affect the success rate of the proposed attack in the setup when the architectures are different (experiment B). However, further increasing the temperature parameter leads to the continuous degradation of the attack. When $\tau = 10.0$, we observe no particular gain compared to the initial experiment ($\tau = 1.0$) in experiment A and observe a significant drop in experiment B. We hypothesize that this happens because of the behaviour of $\mathcal{L}_{\text{KL}}$ when $\tau \rightarrow \infty$:
\begin{equation}
\begin{split}
\label{eq:kl_tau}
\lim_{\tau \rightarrow \infty}\mathcal{L}_{\text{KL}} = \frac{1}{2K}\mathcal{L}_{\text{MSE}} + \delta_{\infty}, 
\end{split}
\end{equation}
where 
\begin{equation}
\label{eq:delta_inf}
\delta_{\infty} = -\frac{1}{2K^2}\left(\sum\limits_{j=1}^K z_j^s - \sum\limits_{j=1}^K z_j^t\right)^2 + c.
\end{equation}
for some constant $c$.
For high values of $\tau$, the task also involves optimizing $\delta_{\infty}$, making shadow models' logits mean to deviate from that of the target's logit mean. This hinders complete logit matching, which is crucial for obtaining reasonably good shadow models. Therefore, following \cite{kim2021comparing}, we propose to employ a Mean-Squared-Loss instead of Kullback-Leibler divergence to ensure direct logit matching when training shadow models. The loss function for training shadow models then follows Eq. \eqref{eq:knowledge_distillation_mse}. As can be seen from Figure \eqref{fig:kl_weight}, we can observe a significant improvement in the success rate of our attack when switching to MSE as a distillation loss in both experiments.

To this end, we propose two versions of $\text{GLiRA}$: the one with the loss function from Eq. \eqref{eq:knowledge_distillation_kl} with $\alpha = 1.0, \ \tau = 1.0$ which we denote $\text{GLiRA}$ $(\text{KL})$, and the one with the loss function from  Eq. \eqref{eq:knowledge_distillation_mse} with $\alpha = 1.0$ which we denote $\text{GLiRA}$ $(\text{MSE})$.

\begin{figure}[ht!]

    \includegraphics[width=\linewidth]{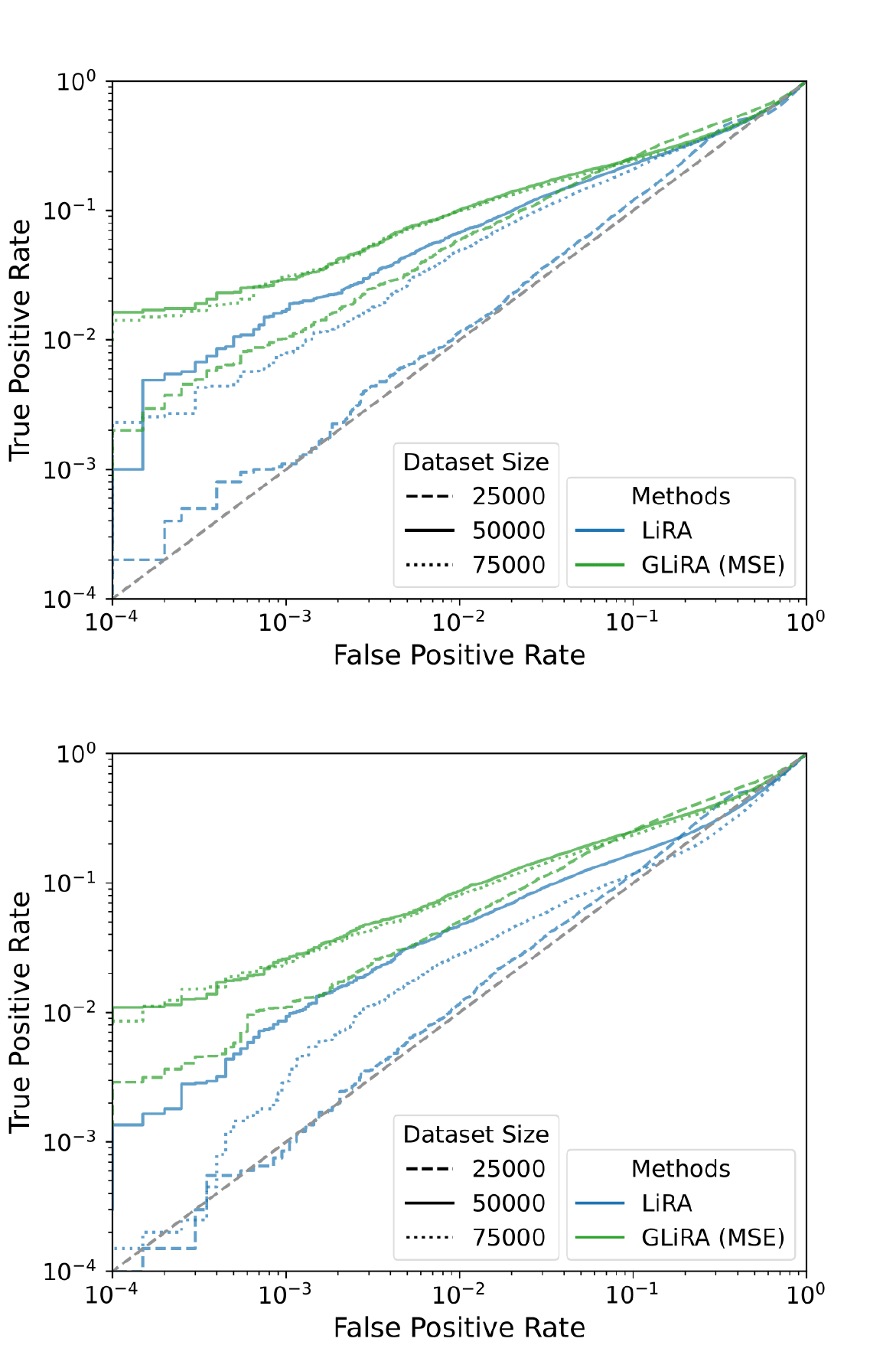}

    \caption{The comparison of the proposed attack approach against LiRA for the different sizes of the training dataset for the shadow models, CINIC10 dataset. \textit{Top}: The architecture of the target model and shadow models is MobileNet-V2; \textit{Bottom}: The architecture of the target model is MobileNet-V2, the architecture of the shadow models is  ResNet34.}
    \label{fig:trainset_size}
\end{figure}

\begin{figure}[ht!]

    \includegraphics[width=1\linewidth]{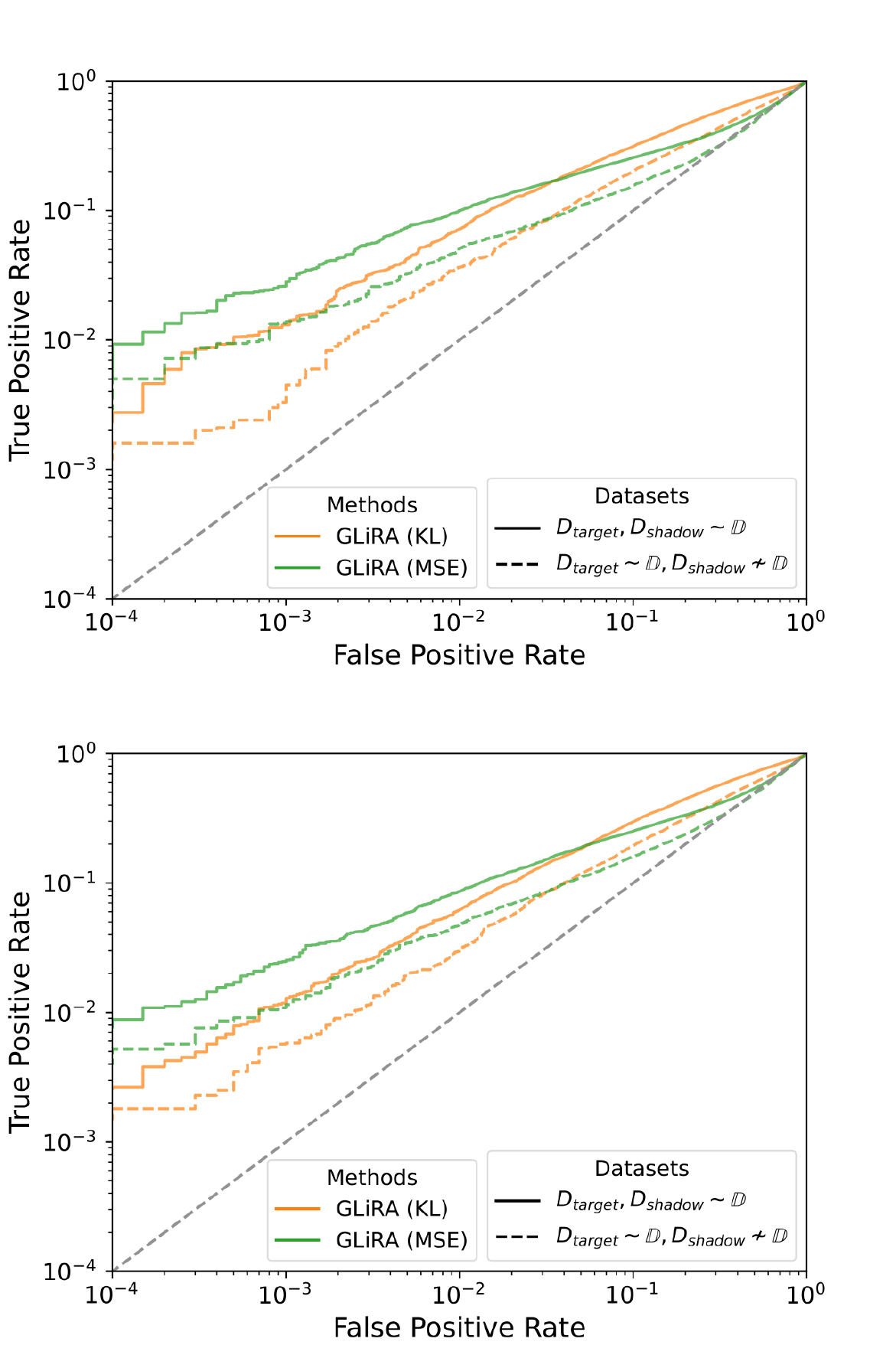}

    \caption{The performance of the proposed attack methods in the setting when training datasets of the shadow models and the training dataset are from the same (solid) and from the different (dashed) distributions. The performance of the method decreases when the distributions differ. \textit{Top}: The architecture of the target and the shadow models is MobileNet-V2; \textit{Bottom}: The architecture of the target model is MobileNet-V2, the architecture of the shadow  models is ResNet34.}
    \label{fig:domain}
\end{figure}

\subsection{Comparison with the state-of-the-art MIA Methods}

It is notable that our method, $\text{GLiRA} \ (\text{KL})$, consistently provides strong performance in the high FPR regime (namely, when FPR $\geq 10.0\%$) by outperforming the concurrent methods in most of the cases. For example, in Table \ref{table:attack_res34_same} we can see that it achieves $0.925$ AUC score, which is almost $10\%$ higher than of \text{Canary} and $15\%$ higher than of the baseline \text{LiRA} approach. In contrast, it can yield unstable performance when examined in the low FPR regime (see Figure \ref{fig:main_same}). For example, when the architecture of the target model and shadow models is the same, $\text{GLiRA} \ (\text{KL})$ outperforms the concurrent approaches on the CIFAR100 dataset (when the architecture is MobileNet-V2); at the same time, it has poor performance on CINIC10 dataset (when the architecture is WideResNet28-10).

On the other hand, $\text{GLiRA} \ (\text{MSE})$ consistently outperforms the concurrent methods on low FPR regime (namely, when FPR $\leq 1.0\%$) and yields comparable attack success in most of the cases for higher FPR, which provides strong evidence of the robustness of our attack compared to other methods. For example, when the architectures of the target model and the shadow ones match, it achieves an improvement of $7\%$ in TPR@$0.1\%$FPR and $16\%$ in TPR@$1\%$FPR on CIFAR100 dataset. When the architectures differ, it achieves an improvement of  $7.5\%$ in TPR@$0.1\%$FPR and $17\%$ in TPR@$1\%$FPR on the CIFAR100 dataset. Remarkably, $\text{GLiRA} \ (\text{MSE})$ outperforms the baseline method, LiRA,  consistently for all values of FPR in all considered experimental setups.

\section{Ablation Study}

\subsection{The Size of Shadow Models' Training Datasets}

The size of the dataset used for knowledge distillation is a crucial parameter for a successful distillation. In this section, we explore the impact of the size of shadow datasets on the performance of the proposed membership inference attack method. To train target and shadow models, we use the architectures from Section \ref{subsec:experiment_kl} and perform experiments on the CINIC10 dataset. Namely, we test the size of $25000, 50000$, and $75000$ while fixing the size of the target model's training dataset at $50000$  samples. We report the results only for one of the proposed methods, namely, for $\text{GLiRA} \ (\text{MSE})$, and compare it against the baseline approach, {LiRA}.


The results are presented in Figure \ref{fig:trainset_size}. As expected, when an adversary is able to acquire a larger dataset for the shadow models' training, the performance of both attacks can be improved. Surprisingly, when the shadow models' dataset size becomes greater than the target model's (namely, $75000$ samples), the performance of the proposed attack method does not change significantly, while the success rate of LiRA notably decreases. We assume that it happens because the shadow models trained on larger datasets tend to generalize better, thereby reducing their effectiveness in mimicking the specific behaviour of the target model. 
Contrarily, our method aims to transfer the behaviour of a given target model explicitly and is resistant to such an issue. 

\subsection{Distribution Shift}

In this section, we test how the performance of the proposed attack method differs when there is a \text{distribution shift} between target and shadow datasets. To model the shift, we assume that an adversary uses an auxiliary dataset to train shadow models. Specifically, we train the target model on the CIFAR10 portion of the CINIC10 dataset, while the shadow models are trained on the ImageNet portion of CINIC10.

The results are reported in Figure \ref{fig:domain}. It is shown that the performance of the proposed approach degrades when there is a distribution shift between the target and shadow models. This may possibly be explained by the difference in the output distributions produced by target model when facing different domains: the shadow models distribution is skewed towards training domain, resulting in the drop in performance during evaluation on the target data domain.

\begin{figure}[h!]
    \includegraphics[width=1\linewidth]{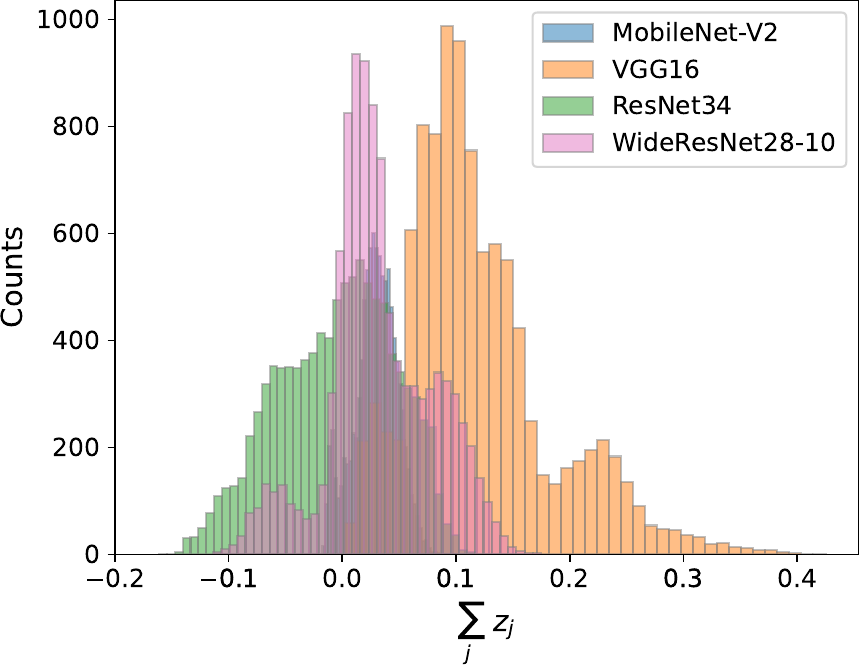}
    \caption{The histograms of the sums of the logits, $\sum_j z_j$. The dataset is  CIFAR10.}
    \label{fig:logit-sum}
\end{figure}

\begin{figure}[h!]
    \includegraphics[width=1\linewidth]{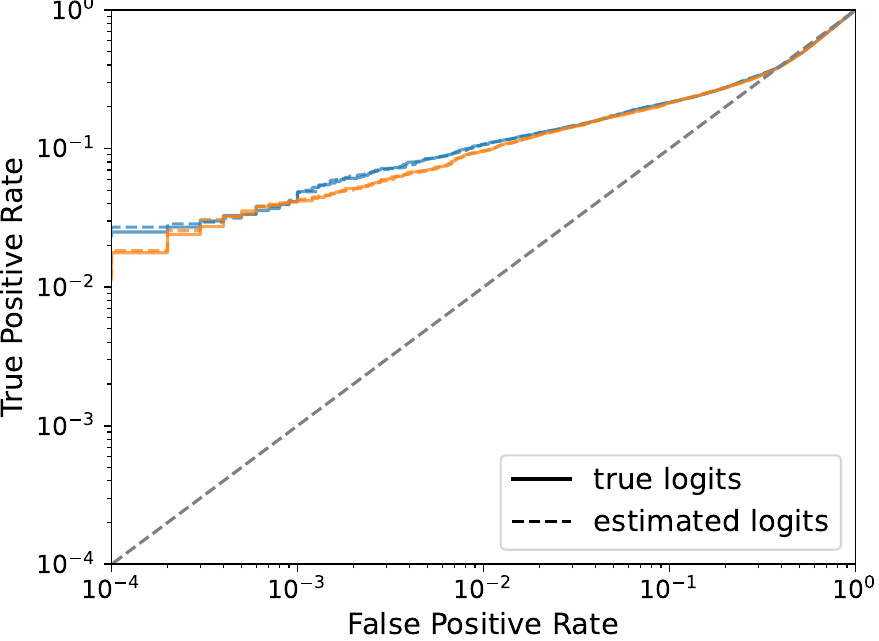}
    \caption{The performance of GLiRA (MSE) when the true logits are replaced by the estimation in the form from Eq. \eqref{eq:black-box-logits}. The dataset is CIFAR10.}
    \label{fig:black-box-mse}
\end{figure}

\subsection{Logits Estimation for MSE}


Note that GLiRA (MSE) leverages information about the logits $z^t$ of the target model (see Eq. \eqref{eq:knowledge_distillation_mse}). In the case when the prediction of the target model in the black-box setting is the vector of probabilities $p^t = \sigma(z^t),$ a potential adversary does not have direct access to the logits. It can be shown that a certain component of the vector of probabilities can be reconstructed up to a constant:


\begin{equation}
    z_k = \ln(p_k) + M, \ M = -\ln\left(\sum\limits_{j=1}^Ke^{z_j}\right),
\end{equation}
where $z_k$ is a logit corresponding to a class $k$. Therefore, to precisely reconstruct logits, additional information about the vector of logits is required.

In \cite{kim2021comparing}, the authors demonstrate that the magnitude of the sum of the logits  $\sum_jz_j$ is close to zero. In their experiments, they evaluate the WideResNet28-4 network trained on the CIFAR100 dataset. Assuming that the sum of the logits is zero, we can find the constant $M$ and, therefore, reconstruct the logits from the probabilities as
\begin{equation}
\label{eq:black-box-logits}
    z_k = \ln(p_k) - \frac{1}{K}\sum\limits_{j=1}^K\ln(p_j).
\end{equation}
In Figure \eqref{fig:logit-sum}, we demonstrate the histograms of the magnitude of the sum of the logits for the target networks trained on the CIFAR10 dataset. 
Indeed, it can be seen that the sum of the logits is very close to zero. Namely, for VGG16, the average sum of the logits among the samples from the dataset is  $0.12$, which corresponds to the largest deviation from zero among all considered architectures. Such a deviation, however, yields a relatively small absolute error ($\approx 0.012$) between the true logit and the reconstructed one in the form from Eq. \eqref{eq:black-box-logits}.

Additionally, in Figure \ref{fig:black-box-mse}, we compare two versions of the method $\text{GLiRA} \ \text{MSE}$. Namely, the solid lines correspond to the setting when an adversary has access to the true logits, and dashed lines correspond to the setting when an adversary uses the reconstructed logits in the form from  Eq. \eqref{eq:black-box-logits}. We observe no significant difference in the success of attack in these two settings. For the first experiment, we used MobileNet-V2 as the architecture of both the target model and shadow models;  for the second experiment, we used MobileNet-V2 as the architecture of the target models and ResNet-34 as the architecture of the shadow models. To this end, if an adversary does not have access to the logits of the target model, we propose to estimate them using Eq. \eqref{eq:black-box-logits}.

\section{Conclusion}
\label{s:conclusion}
In this work, we propose GLiRA, a novel framework that applies knowledge distillation to perform membership inference attacks. We demonstrate that explicit distillation of the target model yields sufficient information to determine the membership status of the input data points. Our approach operates in a black-box setting without requiring any information about the target model. The method can be used to increase the effectiveness of other membership inference attack methods requiring training shadow models. We evaluate our approach on multiple datasets and neural network architectures, comparing it against concurrent methods, and show that our approach outperforms state-of-the-art membership inference attacks in most of the considered experimental setups. Future work includes exploring more fine-grained techniques to transfer target model behaviour and studying ways to reduce computational costs by decreasing the number of trained shadow models.


\bibliographystyle{IEEEtran}  
\bibliography{biblio}  

\begin{IEEEbiography}[{\includegraphics[width=1in,height=1.25in,clip,keepaspectratio]{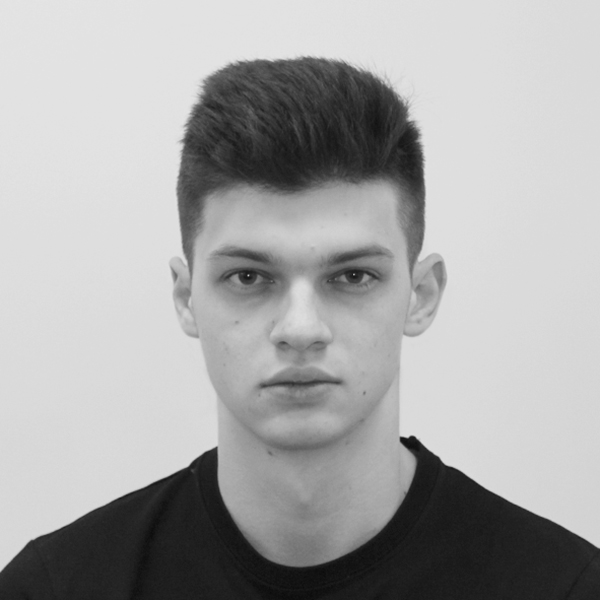}}]{Andrey V. Galichin} received the bachelor’s degree in applied  mathematics from the Lomonosov Moscow State University, in 2022, and the M.Sc. degree in Data Science from the Skolkovo Institute of Science and Technology (Skoltech), Moscow, Russia, in 2024. 
\end{IEEEbiography}

\begin{IEEEbiography}[{\includegraphics[width=1in,height=1.25in,clip,keepaspectratio]{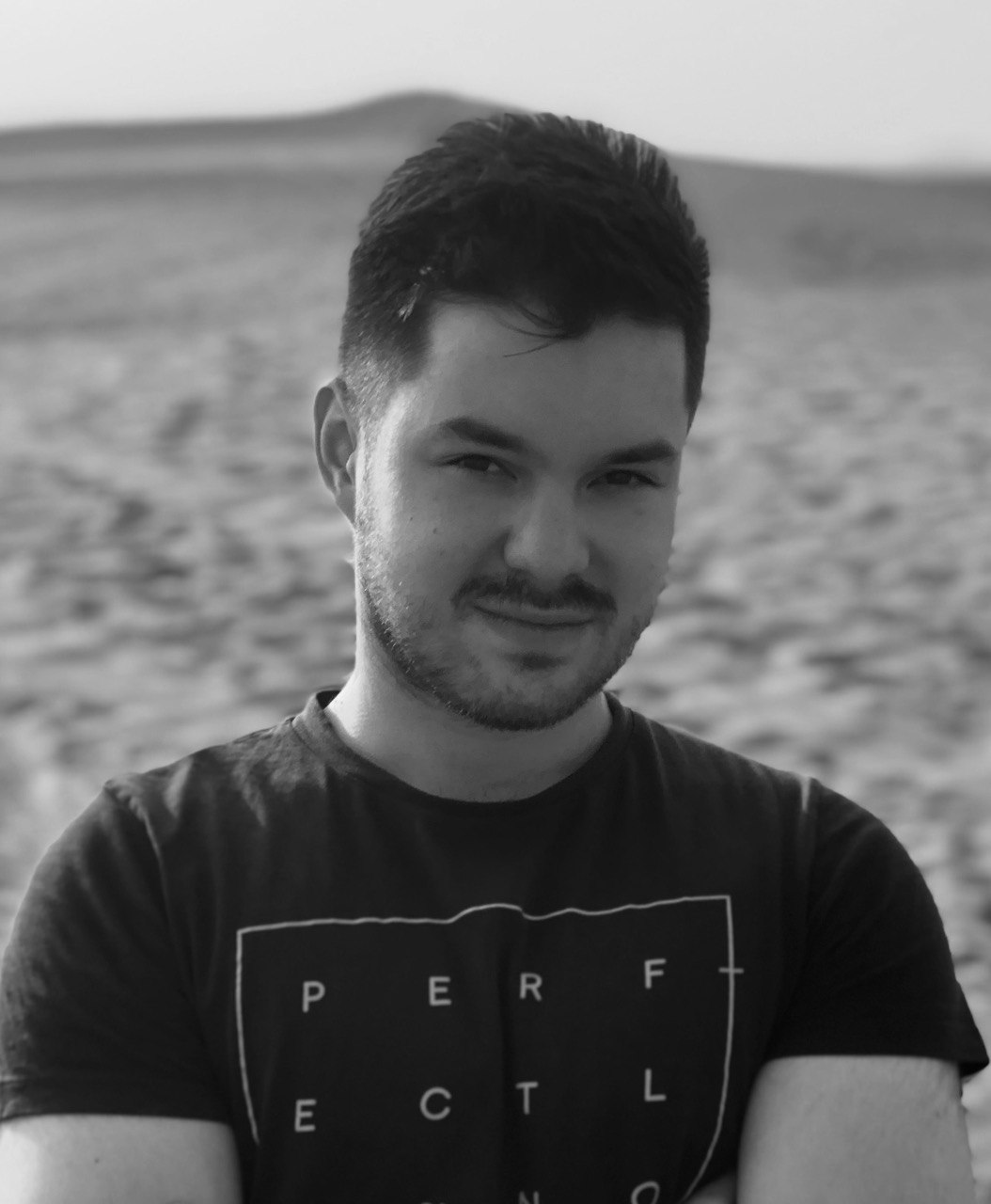}}]{Mikhail Pautov} received the bachelor’s degree in Applied Physics and Mathematics from Moscow Institute of Physics and Technology, Dolgoprudny, Russia, in 2018, and the M.Sc. degree in Data Science from Skolkovo Institute of Science and Technology (Skoltech), Moscow, Russia, in 2020. He finished his Ph.D. studies at Skoltech at 2024. His current research interests include probability theory, certified robustness, and the privacy of neural networks. Dr. Pautov is currently the research scientist at the Reliable and Secure Intelligent Systems Group at the Artificial Intelligence Research Institute (AIRI).
\end{IEEEbiography}

\begin{IEEEbiography}[{\includegraphics[width=1in,height=1.25in,clip,keepaspectratio]{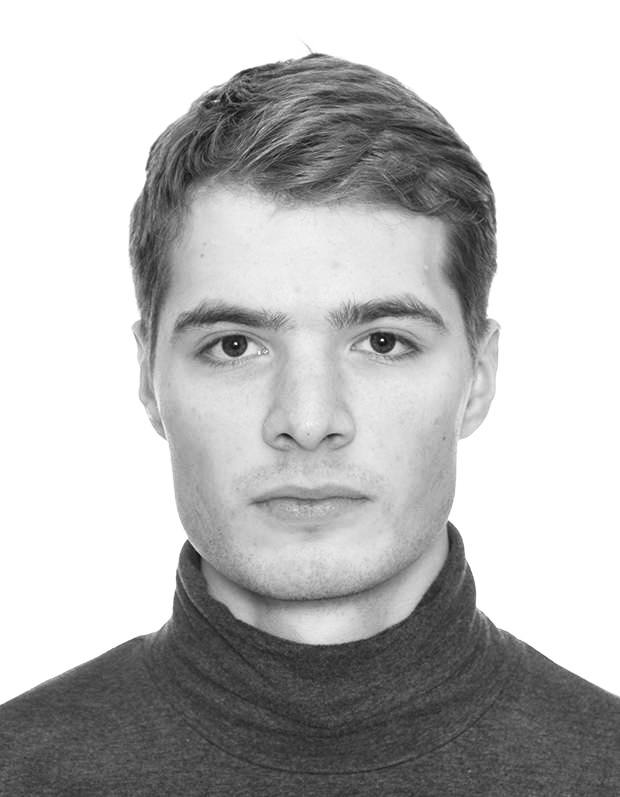}}]{Alexey Zhavoronkin} received the bachelor’s degree and M.Sc. degree in Applied Physics and Mathematics from the Moscow Institute of Physics and Technology, Russia, in 2022 and 2024, respectively. His current research interests include probability theory, generative adversarial networks, and AI safety.
\end{IEEEbiography}

\begin{IEEEbiography}[{\includegraphics[width=1in,height=1.25in,clip,keepaspectratio]{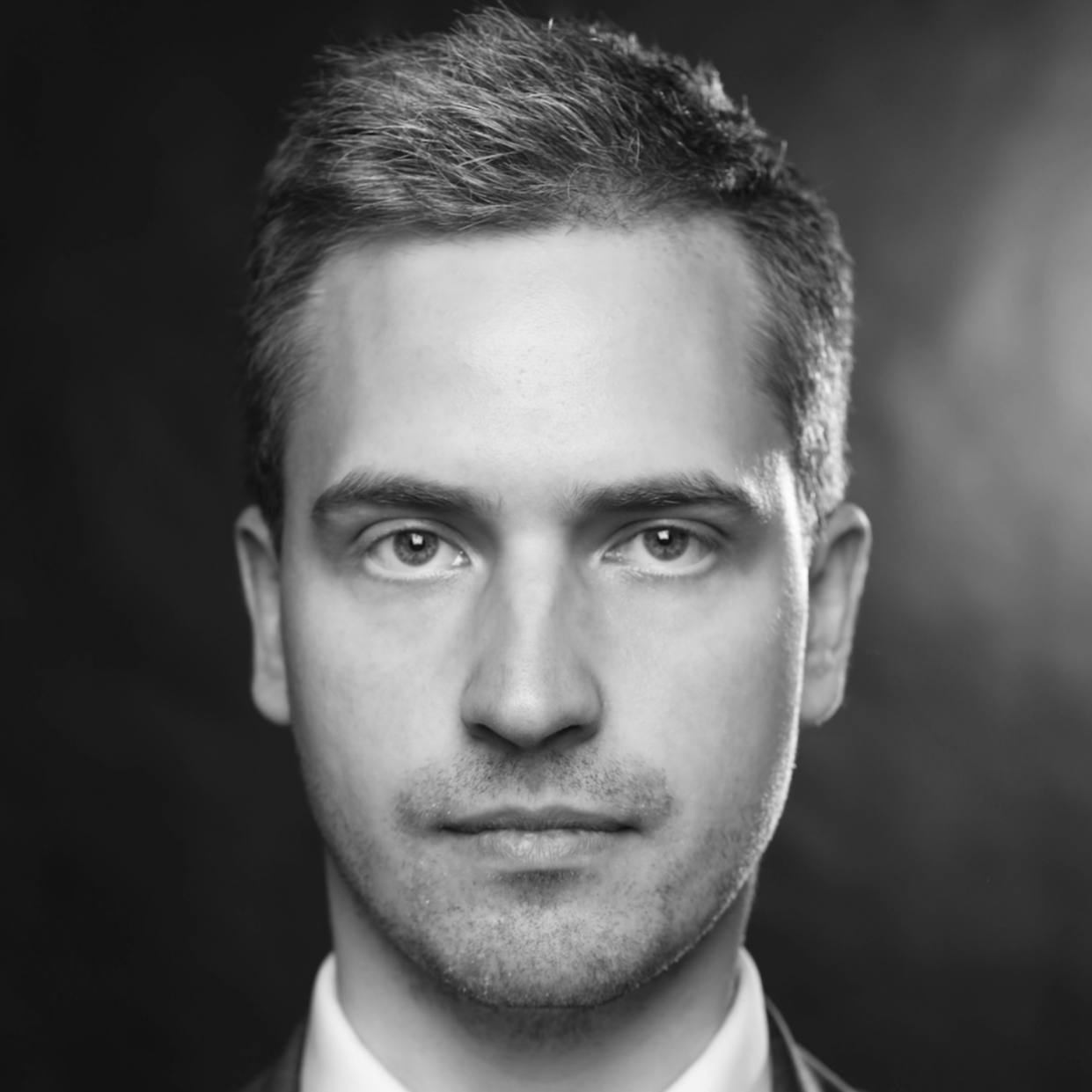}}]{Oleg Y. Rogov} received the M.Sc. degree in physics from the Lomonosov Moscow State University, Moscow, Russia. He received the Ph.D. degree in mathematical modeling and physics at the Russian Academy of Sciences, Moscow. From 2015 to 2019, he was a Research engineer with the Keldysh Institute of  Applied Mathematics, Moscow, Russia. Dr. Rogov is currently the senior research scientist and the head of the Reliable and Secure Intelligent Systems Group at the Artificial Intelligence Research Institute (AIRI). His research interests include AI safety and trustworthiness, data privacy, certified robustness and large language models alignment. 
\end{IEEEbiography}

\begin{IEEEbiography}[{\includegraphics[width=1in,height=1.25in,clip,keepaspectratio]{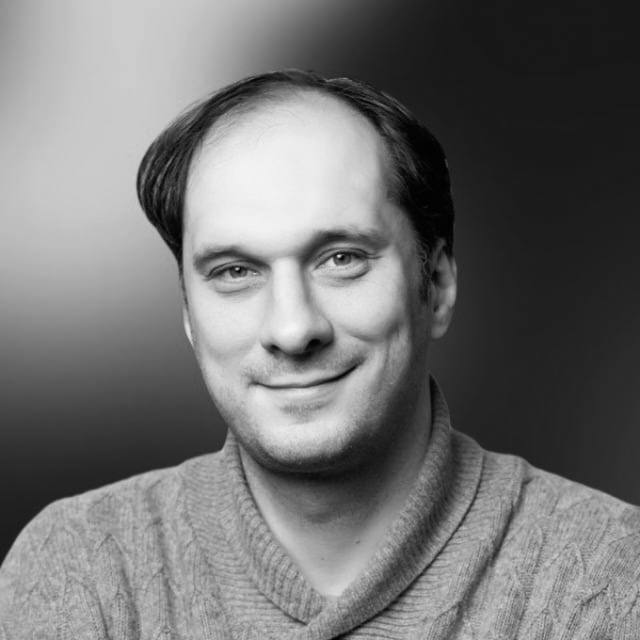}}]{Ivan Oseledets} graduated from the Moscow Institute of Physics and Technology, Dolgoprudny, Russia, in 2006, and received the Ph.D. and D.Sc. degrees from the Institute of Numerical Mathematics of Russian Academy of Sciences (INM RAS) named after G. Marchuk, Moscow, Russia, in 2007 and 2012, respectively. He is currently the CEO of Artificial Intelligence Research Institute and a Leading Researcher with INM RAS. He has coauthored more than 100 papers. His current research interests include linear algebra, tensor methods, machine learning, and deep learning. He is a recipient of several awards, including the SIAM Outstanding Paper Prize in 2018 and the Russian President Award for Science and Innovation for young scientists.
\end{IEEEbiography}


\end{document}